%% file: main.tex
\documentclass{article} % For LaTeX2e
\usepackage{iclr2027_conference,times}

\input{math_commands.tex}

\usepackage[utf8]{inputenc} % allow utf-8 input
\usepackage[T1]{fontenc}    % use 8-bit T1 fonts
\usepackage{xcolor}         % colors (must be before definecolor)
\definecolor{iccvblue}{rgb}{0.21,0.49,0.74}
\definecolor{linkred}{rgb}{1.0,0,0}
\definecolor{refergreen}{rgb}{0,1.0,0}
\usepackage[unicode]{hyperref}       % hyperlinks
\hypersetup{
  colorlinks,
  linkcolor=linkred,
  citecolor=iccvblue
}
\usepackage{url}            % simple URL typesetting
\usepackage{booktabs}       % professional-quality tables
\usepackage{amsfonts}       % blackboard math symbols
\usepackage{fontawesome}    % 特殊图标
\usepackage{wrapfig}        % 文字环绕图/表
\usepackage{nicefrac}       % compact symbols for 1/2, etc.
\usepackage{microtype}      % microtypography
\usepackage{colortbl}  % 用于表格背景
\usepackage{dashrule}  % 虚线表格线（替代 arydshln，不 patch 表格内部，不影响竖线）
\usepackage{newfloat} % 自动定位图表
\usepackage{float} % 自动定位图表
\usepackage{listings}  % 代码高亮现实
\usepackage{multirow}  % 用于表格
\usepackage{booktabs}  % 用于表格
\usepackage{amsmath}  % 增强数学公式排版
\usepackage{bbding}  % 提供一组dingbat字符，如箭头、勾选框等。
\usepackage{mathrsfs}  % 提供一种书法风格的数学字体，用于表示复杂的数学符号
\usepackage{tabularx} % 用于列宽可变长表格
\usepackage{pythonhighlight}  % 提供对 Python 代码的高亮显示支持。
\usepackage{xspace}
\usepackage{enumitem} % 用于 itemize/enumerate 自定义选项
\usepackage{flushend} % 用于balance命令
\lstnewenvironment{PythonB}[1][]{\lstset{style=mypython, frame=none, breaklines=true, #1}}{}
\usepackage[ruled,vlined,linesnumbered]{algorithm2e}
\usepackage{etoolbox}
\makeatletter
\AfterEndEnvironment{algorithm}{\let\@algcomment\relax}
\AtEndEnvironment{algorithm}{\kern1pt\@algcomment}
\let\@algcomment\relax
\newcommand\algcomment[1]{\def\@algcomment{\footnotesize#1}}
\makeatother
\usepackage{soul}  % soul提供高亮功能（例\hl），xcolor已在顶部加载
\sethlcolor{pink}
\usepackage{pifont}
\usepackage{subcaption}  % 支持 subfigure 环境

\usepackage{cleveref}  % 支持 \cref 引用, 必须在 hyperref 之后
\usepackage{tikz}  % 矢量绘图，用于彩色边框 box
\usetikzlibrary{tikzmark}  % 提供 \tikzmarknode，用于画 inline 彩色框

\usepackage{orcidlink}

\makeatletter
\DeclareRobustCommand\onedot{\futurelet\@let@token\@onedot}
\def\@onedot{\ifx\@let@token.\else.\null\fi\xspace}
\def\eg{\emph{e.g}\onedot}

\def\ie{\emph{i.e}\onedot}

\def\vs{\emph{vs}\onedot}
\def\wrt{w.r.t\onedot}

\makeatother
\newcommand{\ourmethod}{{\fontfamily{lmtt}\selectfont\textsc{SEATS}}\xspace}
\newcommand{\trr}{\texttt{TRR}\xspace}
\newcommand{\trrs}{\texttt{TRR}s\xspace}
\newcommand{\prellm}{\texttt{winDivPrune}\xspace}
\newcommand{\wo}{\emph{w/o}\xspace}
\newcommand{\delete}[1]{}
\newcommand{\specialcell}[2][c]{%
  \begin{tabular}[#1]{@{}c@{}}#2\end{tabular}}

\definecolor{cGrey}{HTML}{F3F7F2}
\definecolor{vcolor}{HTML}{4A90D9}
\definecolor{acolor}{HTML}{D94A4A}
\newcommand{\var}[1]{\varinner#1\relax}
\def\varinner#1,#2\relax{\,{\scriptsize(\textcolor{vcolor}{#1}{\text{-}}\textcolor{acolor}{#2})}}
\newcommand{\varapp}[1]{\varappinner#1\relax}
\def\varappinner#1,#2\relax{\textcolor{vcolor}{#1}{\text{-}}\textcolor{acolor}{#2}}
\crefname{table}{Tab.}{Tables}
\crefname{figure}{Fig.}{Figures}
\crefname{section}{Sec.}{Sections}
\crefname{equation}{Eq.}{Equations}
\crefname{appendix}{Appendix}{Appendices}

\newcommand{\graytxt}[1]{\textcolor{gray}{#1}}

\definecolor{boxred}{RGB}{214, 48, 48}
\definecolor{boxorange}{RGB}{255, 153, 51}

\newcommand{\iconimg}{\textcolor{green!60!black}{\faImage}}

\newcommand{\iconaud}{\textcolor[HTML]{F76964}{\faVolumeUp}}
\newcommand{\envelope}{\ding{41}}
\newcommand{\pen}{\ding{45}}
\newcommand{\cmark}{\textcolor{green!60!black}{\ding{51}}}
\newcommand{\xmark}{\textcolor{red}{\ding{55}}}
\newcommand{\modvideo}{\raisebox{-0.3ex}{\includegraphics[width=0.43cm]{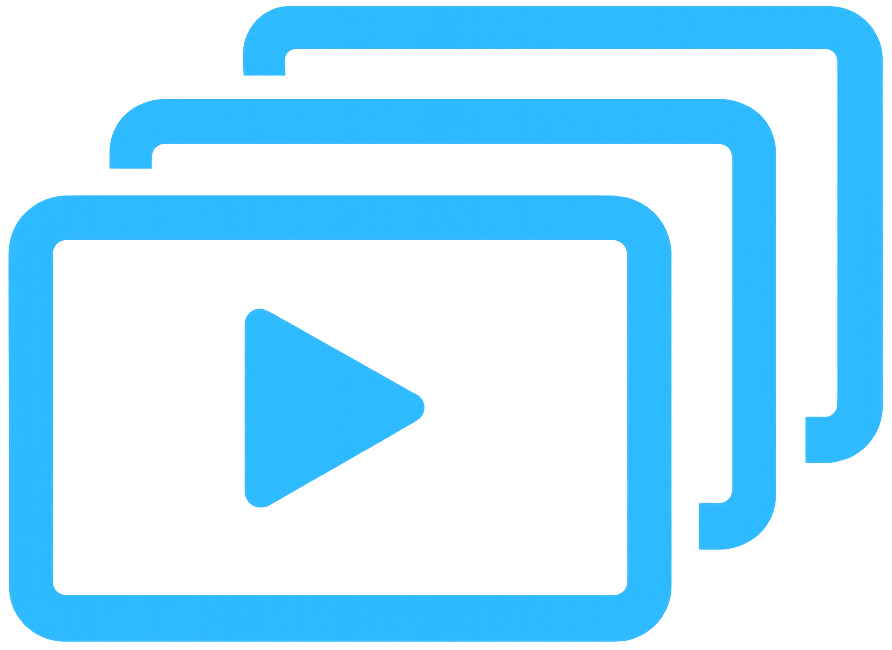}}}

\title{
Stage-adaptive Token Selection for Efficient Omni-modal LLMs
}

\newcommand{\authorname}[2]{\mbox{#1\textsuperscript{\textmd{#2}}}}
\author{%
  \authorname{Zijie Xin}{1}\orcidlink{0000-0002-9220-8735}\quad
  \authorname{Jie Yang}{2,\text{\envelope}}\quad
  \authorname{Ruixiang Zhao}{1}\orcidlink{0009-0008-9984-1841}\quad
  \authorname{Tianyi Wang}{2} \\[1mm]
  \textbf{
    \authorname{Fengyun Rao}{2}\quad
    \authorname{Jing LYU}{2}\quad
    \authorname{Xirong Li}{1,\text{\envelope}}\orcidlink{0000-0002-0220-8310}
  } \\[2mm]
  $^1$~Renmin University of China \quad $^2$~WeChat Vision, Tencent Inc. \\[1mm]
  \small\texttt{\url{https://github.com/xxayt/SEATS}}
}

\iclrfinalcopy % Uncomment for camera-ready version, but NOT for submission.
\begin{document}

\maketitle
\lhead{Under Review}

\renewcommand{\thefootnote}{\pen}
\footnotetext[1]{Work was done when Zijie Xin interned at Tencent. (xinzijie@ruc.edu.cn)}
\renewcommand{\thefootnote}{\envelope}
\footnotetext[2]{Corresponding authors: Xirong Li (xirong@ruc.edu.cn), Jie Yang (cvjieyang@tencent.com)}

{\renewcommand\twocolumn[1][]{#1}
\begin{center}
    \vspace{-6mm}
    \includegraphics[width=0.45\linewidth]{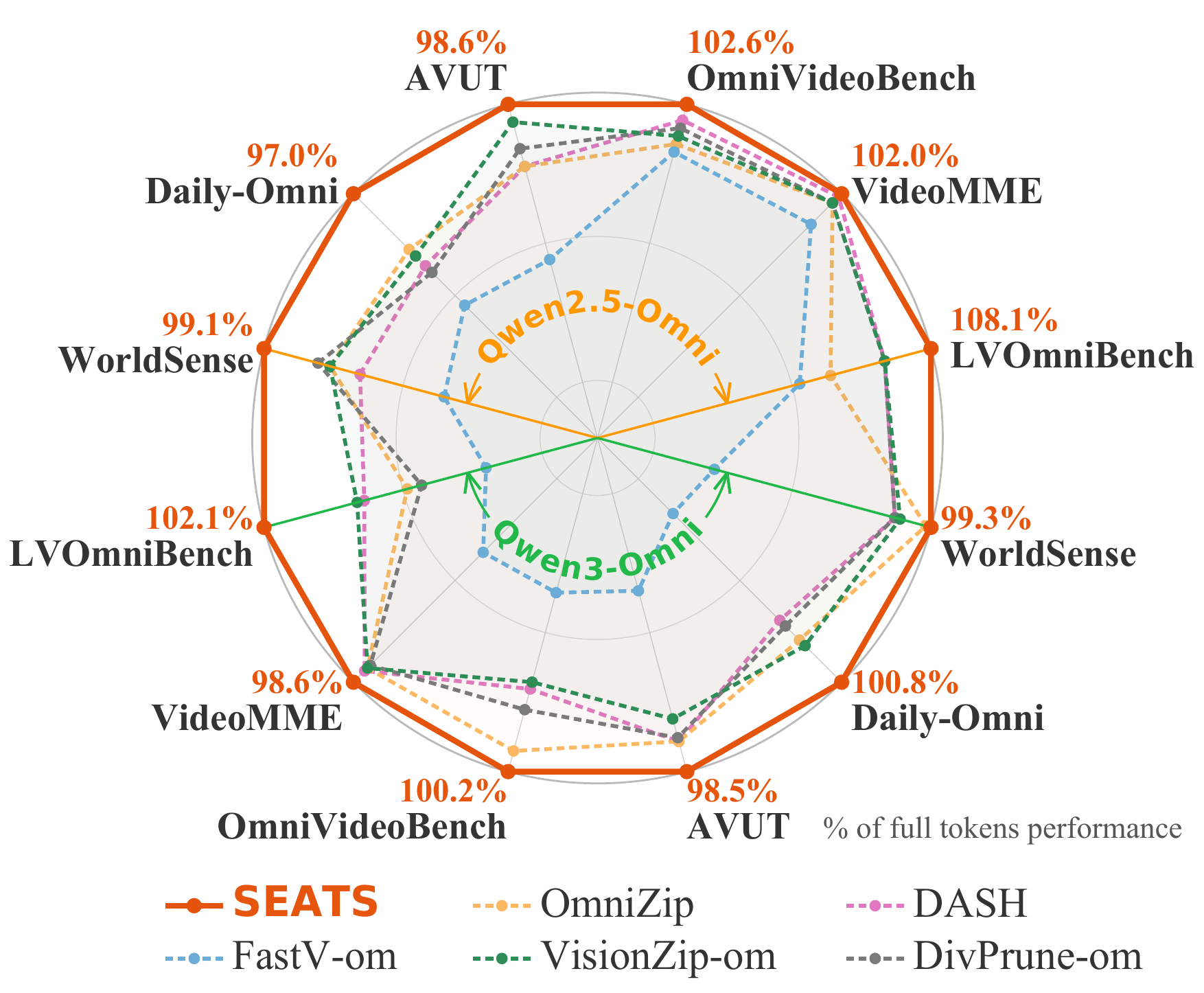}
    \hspace{0.02\linewidth}
    \includegraphics[width=0.42\linewidth]{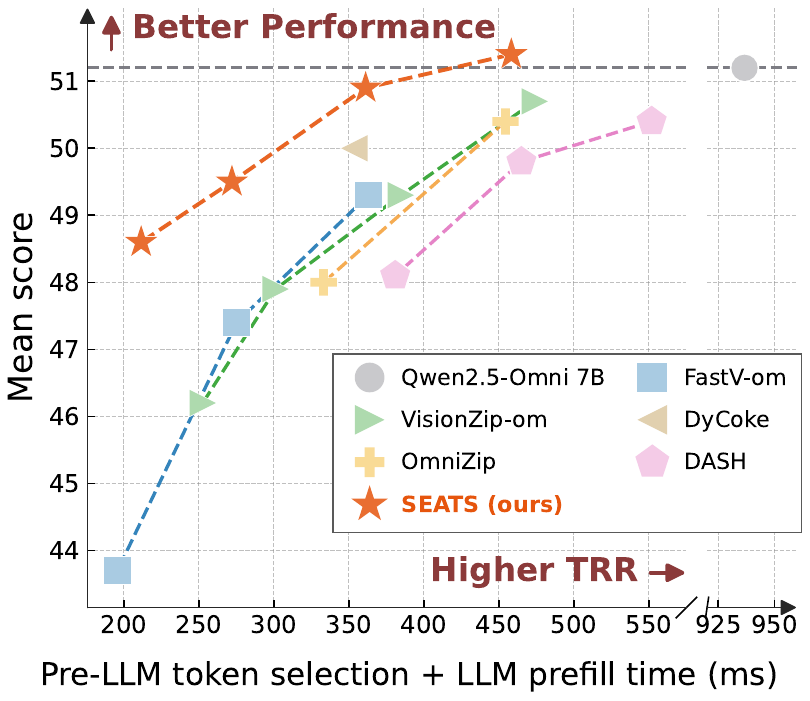}
    \begingroup
    \captionsetup{hypcap=false}
    \captionof{figure}{\textbf{Efficiency--performance trade-off} of training-free token selection methods for om-LLMs.
    }
    \label{fig:teaser}
    \endgroup
\end{center}
}
%-----------------------------------------------------------
\input{sec/0_abstract} \label{sec:abs}
%-----------------------------------------------------------
\section{Introduction} \label{sec:intro}
\input{sec/1_intro}
%-----------------------------------------------------------
\section{Related Work} \label{sec:related}
\input{sec/2_related}
%-----------------------------------------------------------
\section{Token Selection for Om-LLM: Preliminaries and Observations}
\label{sec:analysis}
\input{sec/3_analysis}
%-----------------------------------------------------------
\section{Proposed Method}\label{sec:method}
\input{sec/4_method}
%-----------------------------------------------------------
\section{Experiments} \label{sec:eval}
\input{sec/5_experiment}
%-----------------------------------------------------------
\section{Conclusions} \label{sec:conc}
\input{sec/6_conclusion}

\subsection*{AI Use Statement}

In this work, we used large language models (LLMs) to check for grammar errors, polish the fluency of our writing, and assist in organizing and presenting experimental results. No LLM was used for research ideation, experimental design, or forming conclusions. All substantive intellectual contributions are solely made by the authors. We have reviewed all AI-assisted work and take responsibility for the final content of this work, including text, claims, and artifacts produced with the aid of generative AI.

\subsection*{Ethics Statement}

This work aims to improve the inference efficiency of omni-modal large language models, addressing a key barrier to deployment and scalability. By reducing computational cost and memory consumption, it broadens access to advanced audio-visual AI in resource-constrained settings. All benchmarks used in our experiments are publicly available. We do not foresee negative societal impacts beyond those inherent to the underlying om-LLMs.

\subsection*{Reproducibility Statement}

To facilitate reproducibility, we provide evaluation benchmark details in Appendix~\ref{app:benchmark_details}, implementation and configuration of \ourmethod in Appendix~\ref{app:method_details}, and baseline reproduction details in Appendix~\ref{app:baseline_details}.
Our source code and evaluation scripts are included in the supplementary materials.

\subsection*{Acknowledgments}
This research was supported by NSFC (No.62576348), BJNSF (No.L254039), Tencent WeChat Rhino-Bird Focused Research Program, and the Outstanding Innovative Talents Cultivation Funded Programs 2025 of Renmin University of China.

\bibliography{main}
\bibliographystyle{iclr2027_conference}

%%%%%%%%%%%%%%%%%%%%%%%%%%%%%%%%%%%%%%%%%%%%%%%%%%%%%%%%%%%%
\clearpage
\renewcommand{\thefootnote}{\arabic{footnote}}
\setcounter{footnote}{0}
\input{sec/7_appendix}

\end{document}

%% file: math_commands.tex
\usepackage{amsmath,amsfonts,bm}

\def\eqref#1{equation~\ref{#1}}
\def\1{\bm{1}}

\def\vs{{\bm{s}}}

\DeclareMathAlphabet{\mathsfit}{\encodingdefault}{\sfdefault}{m}{sl}
\SetMathAlphabet{\mathsfit}{bold}{\encodingdefault}{\sfdefault}{bx}{n}

%% file: sec/0_abstract.tex
\begin{abstract}

The unified audio-visual understanding capability of omni-modal large language models (om-LLMs) comes at the high cost of processing tens of thousands of visual and audio tokens throughout the LLM. To reduce the cost, training-free methods for om-LLM token selection are being actively developed, operating either pre-LLM or inner-LLM. Recognizing that the non-textual token redundancy varies across stages and LLM layers, we propose in this paper a novel \underline{s}tag\underline{e}-\underline{a}daptive \underline{t}oken \underline{s}election (\ourmethod) method that operates in both pre-LLM and inner-LLM stages. Given an overall token budget measured by the token retention rate (TRR), \ourmethod performs top-down token budget allocation in an input-dependent manner. Specifically, a larger TRR is used for pre-LLM token selection, whilst within the LLM, all tokens are retained in shallow layers, increasingly decreased TRRs are applied in middle layers, and the non-textual tokens are fully removed in late layers. Extensive experiments on three representative om-LLMs and six benchmarks verify the viability of the proposed method. By retaining only 10\% of visual and audio tokens, \ourmethod preserves 94.9\% of the full-token performance on Qwen2.5-Omni while achieving a $9.3\times$ FLOPs reduction and a $4.9\times$ prefill speedup.
\end{abstract}

%% file: sec/1_intro.tex
Omni-modal large language models (om-LLMs) have shown great potential for unified audio-visual understanding \citep{team2026qwen35omni,ye2025omnivinci,li2025baichuanomni15,tang2025videosalmonn2}.
They encode video frames and audio tracks into temporally aligned token sequences and concatenate them with text tokens for joint LLM reasoning.
Their performance relies on dense video frame sampling and high-resolution audio encoding, typically yielding tens of thousands of visual and audio tokens to be fed into an LLM. 
Processing such lengthy tokens throughout the LLM incurs substantial computation and memory overhead. 
Therefore, selecting compact yet semantically sufficient visual and audio tokens is crucial for efficient om-LLM inference.

Token selection~\citep{shao2025compressionsurvey,yao2026compressionsurvey} has been widely studied for image-LLMs~\citep{li2024llavaonevision,li2024llavanextinterleave} and video-LLMs~\citep{zhang2025llavavideo,wang2025internvl35,bai2025qwen3vl,yu2025minicpmv45,ge2025archunyuanvideo7b}.
Depending on where selection is performed, existing methods can be broadly categorized into pre-LLM methods and inner-LLM methods. 
Pre-LLM methods~\citep{yang2025visionzip,alvar2025divprune,deng2025scope} reduce input length using encoder-side signals before LLM computation, but are often query-agnostic and may discard task-critical tokens. 
Inner-LLM methods~\citep{chen2024fastv,wu2026hidrop,xing2024pyramiddrop} exploit text-to-vision attention for query-aware pruning, but shallow-layer attention is noisy, while late pruning limits computational savings. 
For video-LLMs, spatiotemporal redundancy further motivates frame-aware selection~\citep{shenfastvid,fan2026flashvid} and hybrid pre-/inner-LLM strategies~\citep{shao2025holitom,tao2025dycoke,du2026unist}. 
Despite these advances, existing methods are mainly developed for image- and video-LLMs, so they naturally fall short when applied to the temporally interleaved audio-visual tokens of om-LLMs.

Recent studies have begun to explore token selection for om-LLMs, the majority of which operate in the pre-LLM stage.  
OmniZip~\citep{tao2025omnizip} and DASH~\citep{li2025dash}, for instance, use audio encoder attention to guide video token pruning.  EchoingPixels pools audio and video tokens for cross-modal joint filtering \citep{gong2025echoingpixels}. 
OmniSIFT performs spatiotemporal video pruning followed by visual-semantic-guided audio token selection \citep{ding2026omnisift}. 
By contrast, OmniFit performs token reduction inside an om-LLM \citep{wang2026omnifit}, but lacks a unified and adaptive token-retention budget allocation strategy across LLM layers, temporal windows, and modalities.
Our empirical analysis reveals a clear block-wise dependence pattern: LLM layers in \emph{shallow} blocks strongly rely on non-textual tokens for cross-modal fusion, layers in \emph{middle} blocks gradually reduce this dependence, while for \emph{late} blocks where the fusion is largely completed, the non-textual tokens are dispensable. 
We note that the design of existing methods actually runs counter to this pattern: FastV prunes visual tokens in shallow layers, where such tokens remain crucial for cross-modal fusion \citep{chen2024fastv}, while HiDrop~\citep{wu2026hidrop} retains more visual tokens in deeper layers, where these tokens have  already become redundant. 
This motivates us to search for a stage-adaptive, depth-aware, and modality-flexible token selection strategy for om-LLMs.

Designing such a strategy is non-trivial due to challenges as follows. 
First, the nature of token redundancy differs across stages: pre-LLM tokens mainly exhibit spatiotemporal similarity, whereas inner-LLM tokens become query-aligned and should be selected in an \emph{input-dependent} manner. 
Second, reliance on non-textual tokens decreases with the LLM's layer depth, making a uniform pruning ratio either too aggressive for shallow layers or too conservative for deeper layers. 
Third, audio-visual importance varies across temporal windows, where either modality may provide the key evidence. 
Such varied importance cannot be handled by a fixed token retention ratio (\trr).

To address these challenges, we propose \ourmethod, a training-free \underline{S}tag\underline{E}-\underline{A}daptive \underline{T}oken \underline{S}election method for efficient om-LLM inference. 
Before the LLM, \ourmethod applies attention-weighted diversity selection within each temporal window to remove spatiotemporal redundancy and shorten the input sequence. 
Inside the LLM, it adopts a model-agnostic block-wise token-retention-ratio (\trr) decay schedule, progressively increasing pruning strength as the dependence on non-textual tokens decreases. 
It further distributes the retention budget through a top-down two-level allocation strategy, first across temporal windows and then across modalities, guided by query relevance scores. 
In late layers, where cross-modal fusion is largely completed, \ourmethod removes all remaining non-textual tokens so that subsequent layers process only text tokens. 
Together, these stages enable token selection that adapts to both layer-wise dependency and cross-modal dynamics without retraining.

Extensive experiments on six audio-visual benchmarks and three representative om-LLMs, Qwen2.5-Omni-7B, OmniVinci-9B and Qwen3-Omni-30B, verify the viability of \ourmethod.
It is comparable to the full-token performance while using only 33\% computational cost on Qwen2.5-Omni-7B, see~\cref{fig:teaser}.  
At a \trr of 0.1, it achieves a $9.3\times$ FLOPs reduction and a $4.9\times$ prefill speedup while preserving 94.9\% of the original performance. To sum up, our main contributions are as follows: \\
$\bullet$ \textbf{Insight}. 
We reveal a block-wise modality dependence pattern in current om-LLMs (OmniVinci and Qwen-Omni series), where visual and audio tokens become less important for deeper LLM layers. This finding sheds light on a more aggressive yet effective path to om-LLM token selection.
\\
$\bullet$ \textbf{Solution}.
The proposed \ourmethod operates in both pre-LLM and inner-LLM stages. For pre-LLM token selection, we extend DivPrune \citep{alvar2025divprune} to the omni-modal context and in a window-based manner. Our inner-LLM  token selection follows a block-wise \trr decay schedule and query-guided top-down token budget allocation. \\
$\bullet$ \textbf{Results}. 
Experiments on three om-LLMs and six benchmarks show that \ourmethod achieves a superior efficiency-performance trade-off,  compared with a number of strong \emph{open-source} baselines including FastV \citep{chen2024fastv}, VisionZip~\citep{yang2025visionzip}, and DivPrune~\citep{alvar2025divprune} originally for image-LLMs, DyCoke~\citep{tao2025dycoke} and FastVID~\citep{shenfastvid} developed for video-LLMs, and OmniZip~\citep{tao2025omnizip} and DASH~\citep{li2025dash} dedicated to om-LLMs.

%% file: sec/2_related.tex
As we aim for training-free token selection, we discuss recent progress in this line of research (\cref{table:related}).

\input{table/table_related_v6}

\textbf{For image-LLMs}.
Depending on whether token selection is performed before or inside the LLM, existing methods can be divided into two groups: pre-LLM \citep{yang2025visionzip,shang2024llavaprumerge,zhang2025vispruner,deng2025scope,zhang2025cdpruner,dong2026mmtok} and inner-LLM \citep{chen2024fastv,xing2024pyramiddrop,zhang2024sparsevlm,wu2026hidrop}.
For pre-LLM token selection, VisionZip~\citep{yang2025visionzip}, LLaVA-PruMerge~\citep{shang2024llavaprumerge}, and VisPruner~\citep{zhang2025vispruner} measure token saliency via \texttt{[CLS]} attention. DivPrune \citep{alvar2025divprune} formulates token selection as a max-min diversity problem. SCOPE~\citep{deng2025scope} and CDPruner~\citep{zhang2025cdpruner}  consider both saliency and diversity, whilst MMTok~\citep{dong2026mmtok} performs multimodal coverage-based selection. Since visual and textual tokens are not semantically aligned in the pre-LLM stage, these methods  are typically user-query agnostic. By contrast, inner-LLM methods prune visual tokens at specific LLM layers based on text-to-vision attention, making them inherently query-aware. FastV~\citep{chen2024fastv} performs one-shot pruning at a shallow layer.  PDrop~\citep{xing2024pyramiddrop} and SparseVLM~\citep{zhang2024sparsevlm} perform token selection across multiple layers with a fixed \trr.
HiDrop~\citep{wu2026hidrop} operates at middle-to-deep layers with a concave schedule such that deeper layers are assigned larger \trrs. Different from HiDrop, \ourmethod employs a model-agnostic stage-adaptive \trr decay schedule, where \trr progressively decreases as LLM layers go deep.

\textbf{For video-LLMs}. Pre-LLM methods have been extended to the video domain by exploiting inter-frame token redundancy, see for instance FastVID \citep{shenfastvid},
FlashVID \citep{fan2026flashvid}, and VidCom2 \citep{liu2025vidcom2}. Meanwhile, we observe a growing interest in jointly using pre-LLM and inner-LLM approaches \citep{tao2025dycoke,shao2025holitom,huang2025prunevid}. DyCoke first merges temporally redundant tokens in the pre-LLM stage, and then dynamically reduces the KV cache within the LLM  \citep{tao2025dycoke}. HoliTom performs both pre-LLM and inner-LLM token merging \citep{shao2025holitom}.  PruneVID \citep{huang2025prunevid} and UniST \citep{du2026unist} first perform spatial-temporal merging in the pre-LLM stage, and then conduct query-aware token selection inside the LLM. As these methods are designed for single-modality (visual) token pruning, directly applying them to om-LLMs, 
say by handling the visual and audio tokens in parallel, is suboptimal.

\textbf{For om-LLMs}. Token selection for om-LLMs is mostly conducted in the pre-LLM stage. 
OmniZip uses audio saliency to guide video pruning in fixed-length windows \citep{tao2025omnizip}.
DASH replaces the windows  by variable-length audio-semantic segments \citep{li2025dash}.
OmniScope resorts to multiple external models for token-to-query relevance estimation, thus incurring extra computation overhead \citep{su2026omniscope}. In contrast to the above works, OmniFit performs layer-wise token selection within the LLM \citep{wang2026omnifit}.  Recognizing that token redundancy grows with layer depth, OmniFit allocates smaller token retention budgets to deeper layers. However, its budget allocation scheme is computed offline and thus cannot adapt to individual inputs.

%% file: table/table_related_v6.tex
\begin{table}[htbp!]
\caption{
\textbf{Training-free token selection for image-(\iconimg), video-(\modvideo), and om-(\modvideo+\iconaud) LLMs }. 
}
\label{table:related}

\centering
\setlength{\tabcolsep}{3.0pt}
\renewcommand{\arraystretch}{1.05}

\resizebox{0.98\linewidth}{!}{
\begin{tabular}{@{}lcc c cc c cc c c@{}}
\toprule

\multirow{2}{*}{\textbf{Method}} & 
\multirow{2}{*}{\textbf{\specialcell{Open-source}}} & 
\multirow{2}{*}{\textbf{\specialcell{External-model\\Free}}} &
\multirow{2}{*}{\textbf{\specialcell{Targeted\\MLLM}}}
& \multicolumn{2}{c}{\textbf{Selection Stage}}
&
& \multicolumn{2}{c}{\textbf{Selection Criterion}}
& \multirow{2}{*}{\textbf{Adaptive \trr}}
& \multirow{2}{*}{\textbf{\specialcell{Late\\Removal}}} \\
\cmidrule{5-6} \cmidrule{8-9}

& & & & \textit{Pre-LLM} & \textit{Inner-LLM} & &
\textit{Relevance} & \textit{Diversity} & & \\
\midrule

%% ============ For ImageLLMs ============
VisionZip~\citep{yang2025visionzip} \tiny{CVPR'25}
& \cmark & \cmark & \iconimg & \cmark & \xmark & & \iconimg & \xmark & \xmark & \xmark \\

DivPrune~\citep{alvar2025divprune} \tiny{CVPR'25}
& \cmark & \cmark & \iconimg & \cmark & \xmark & & \xmark & \iconimg & \xmark & \xmark \\

CDPruner~\citep{zhang2025cdpruner} \tiny{NIPS'25}
& \cmark & \xmark & \iconimg & \cmark & \xmark & & \xmark & \iconimg & \xmark & \xmark \\

SCOPE~\citep{deng2025scope} \tiny{NIPS'25}
& \cmark & \cmark & \iconimg & \cmark & \xmark & & \iconimg & \iconimg & \xmark & \xmark \\

FastV~\citep{chen2024fastv} \tiny{ECCV'24}
& \cmark & \cmark & \iconimg & \xmark & \cmark & & \iconimg & \xmark & \xmark & \xmark \\

PDrop~\citep{xing2024pyramiddrop} \tiny{CVPR'25}
& \cmark & \cmark & \iconimg & \xmark & \cmark & & \iconimg & \xmark & \cmark & \xmark \\

HiDrop~\citep{wu2026hidrop} \tiny{ICLR'26}
& \cmark & \cmark & \iconimg & \xmark & \cmark & & \iconimg & \xmark & \cmark & \cmark \\ [2pt]

%% ============ For VideoLLMs ============
FastVID~\citep{shenfastvid} \tiny{NIPS'25}
& \cmark & \cmark & \modvideo & \cmark & \xmark & & \modvideo & \modvideo & \xmark & \xmark \\

DyCoke~\citep{tao2025dycoke} \tiny{CVPR'25}
& \cmark & \cmark & \modvideo & \cmark & \cmark & & \modvideo & \modvideo & \xmark & \xmark \\

HoliTom~\citep{shao2025holitom} \tiny{NIPS'25}
& \cmark & \cmark & \modvideo & \cmark & \cmark & & \modvideo & \modvideo & \xmark & \xmark \\

FlashVID~\citep{fan2026flashvid} \tiny{ICLR'26}
& \cmark & \cmark & \modvideo & \cmark & \cmark & & \modvideo & \modvideo & \xmark & \xmark \\

UniST~\citep{du2026unist} \tiny{CVPR'26}
& \xmark & \cmark & \modvideo & \cmark & \cmark & & \modvideo & \modvideo & \xmark & \xmark \\ [2pt]

% \midrule

%% ============ For OmniLLMs ============
OmniZip~\citep{tao2025omnizip} \tiny{CVPR'26}
& \cmark & \cmark & \modvideo+\iconaud & \cmark & \xmark & & \iconaud & \modvideo & \xmark & \xmark \\

DASH~\citep{li2025dash} \tiny{ECCV'26}
& \cmark & \cmark & \modvideo+\iconaud & \cmark & \xmark & & \iconaud & \modvideo+\iconaud & \xmark & \xmark \\

OmniScope~\citep{su2026omniscope} \tiny{MM'26}
& \cmark & \xmark & \modvideo+\iconaud & \cmark & \xmark & & \modvideo+\iconaud & \modvideo+\iconaud & \xmark & \xmark \\

OmniFit~\citep{wang2026omnifit} \tiny{ICML'26}
& \xmark & \cmark & \modvideo+\iconaud & \xmark & \cmark & & \xmark & \modvideo+\iconaud & \xmark & \xmark \\  [2pt]

\rowcolor[HTML]{f0edff}
\ourmethod~(\emph{ours})
& \cmark & \cmark & \modvideo+\iconaud & \cmark & \cmark & & \modvideo+\iconaud & \modvideo+\iconaud & \cmark & \cmark \\

\bottomrule
\end{tabular}
}

\end{table}

%% file: sec/3_analysis.tex
\input{figure/early_remove_worldsense}

%+++++++++++++
\subsection{Preliminaries}
\label{ssec:analysis_preliminary}
%+++++++++++++

Let $\mathcal{V}$ be a specific video accompanied with an audio track $\mathcal{A}$. Given a user-provided prompt query $\mathcal{Q}$, an om-LLM answers with respect to the video by first encoding the video content as a sequence of $N_v$ visual tokens, the audio track as a sequence of $N_a$ audio tokens and the query as a sequence of $N_q$ textual tokens. Each token is a $d$-dimensional vector, denoted by $z$. When necessary, we use $z_v$, $z_a$ and $z_q$ to denote visual, audio and textual tokens, respectively. These token sequences are then concatenated and fed into an $\mathrm{L}$-layer LLM, which generates a response to the query by producing a new sequence of textual tokens in an autoregressive manner.

For temporal alignment between the visual and audio modalities, the visual and audio token sequences are first partitioned using a fixed-size window, resulting in $T$ non-overlapping windows. For each window $t$, the visual and audio tokens that fall within it are grouped as $[z^{(t)}_{v,1},\ldots,z^{(t)}_{v,n_v},z^{(t)}_{a,1},\ldots,z^{(t)}_{a,n_a}]$, where $n_v$ and $n_a$ indicate the number of visual and audio tokens in that window, respectively. These $T$ groups are then concatenated in chronological order, followed by the textual tokens, to form the input sequence of length $N_v+N_a+N_q$ to the LLM. Since $N_o=N_v+N_a  \gg N_q$, token selection for efficient LLM prefill effectively reduces to selecting  the visual and audio tokens only, with the textual tokens kept entirely intact.

For each layer $l$ in the LLM, let $r_l$ be the token retention ratio (\trr) applied to its input, which reduces the input length from $N_o+N_q$ to $N_o \cdot r_l+N_q$. The value of $r_l$ governs the trade-off between model performance and efficiency. Intuitively, $r_l$ needs to be proportional to the importance of layer $l$. Given the overall \trr $R$ as a token-budget indicator, \ie  $\sum_{l=1}^{\mathrm{L}} r_l \le \mathrm{L} \cdot R$, more important layers should be assigned larger $r_l$ values.  Meanwhile, given $R_v$ and $R_a$ as the overall \trr for visual and audio tokens, respectively, we  have $N_o \cdot R = N_v \cdot R_v + N_a \cdot R_a$.

%+++++++++++++
\subsection{Observations}
\label{ssec:analysis_early_remove}
%+++++++++++++

To empirically identify layer importance, we examine the effect of removing all visual and/or audio tokens at a specific LLM layer of an om-LLM. This approach allows us to measure the extent to which each layer relies on these non-textual tokens. 

As shown in~\cref{fig:early_remove_worldsense}, a consistent trend emerges across three representative om-LLMs (Qwen2.5-Omni-7B \citep{xu2025qwen25omni}, OmniVinci-9B \citep{ye2025omnivinci}, and Qwen3-Omni-30B \citep{xu2025qwen3omni}). 
When $l$ falls within the first 50\% of layers, which we term the \emph{shallow} block, removal causes a clear performance collapse, indicating that the visual and audio information has not yet been absorbed by the textual tokens.
As $l$ goes beyond 50\%, \ie into the \emph{middle} block, model performance recovers rapidly, suggesting that intensive cross-modal fusion is underway and the textual tokens are progressively acquiring the needed audio-visual semantics.
Once $l$ exceeds roughly 80\% of the total depth, entering the \emph{late} block, removal causes almost no performance drop, indicating that the non-textual tokens are no longer needed.

The above results reveal a clear block-wise pattern of layer importance. 
Layers in the shallow block critically depend on the visual and audio tokens and thus demand a relatively high \trr. By contrast, layers in the middle block are more resistant to token removal as cross-modal fusion proceeds, so they can be allocated smaller \trr values. As for the late-block layers, the non-textual tokens can be safely removed without affecting model performance.

%% file: figure/early_remove_worldsense.tex
\begin{figure}[tb]
    \centering
    \begin{subfigure}[t]{0.324\linewidth}
        \centering
        \includegraphics[width=\linewidth]{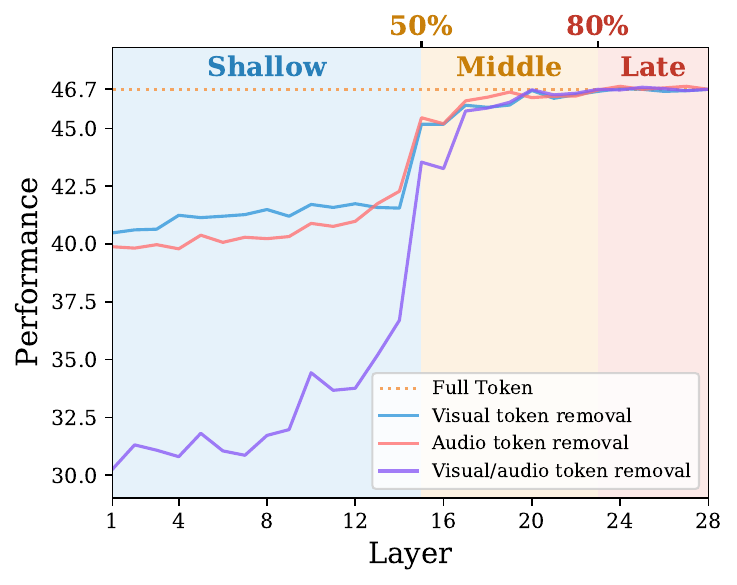}
        \caption{Qwen2.5-Omni-7B}
        \label{fig:early_remove_performance_qw25o}
    \end{subfigure}
    % \hfill
    \hspace{0.001\linewidth}
    \begin{subfigure}[t]{0.324\linewidth}
        \centering
        \includegraphics[width=\linewidth]{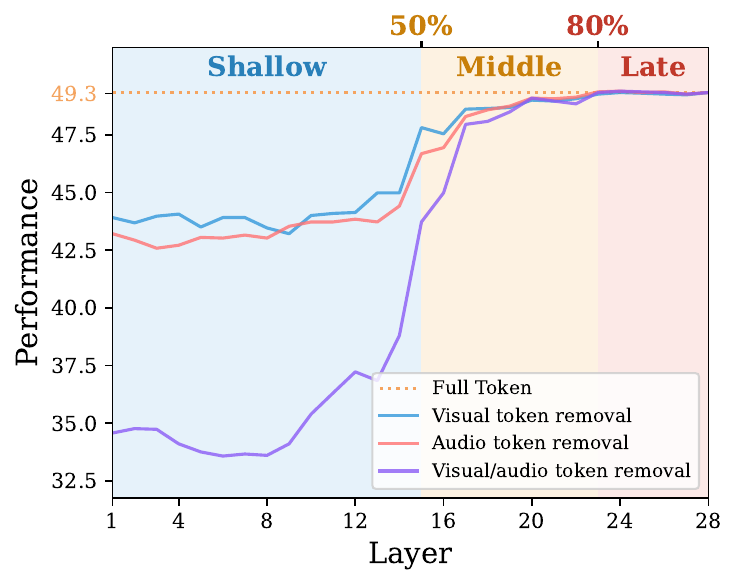}
        \caption{OmniVinci-9B}
        \label{fig:early_remove_performance_omnivinci}
    \end{subfigure}
    % \hfill
    \hspace{0.001\linewidth}
    \begin{subfigure}[t]{0.324\linewidth}
        \centering
        \includegraphics[width=\linewidth]{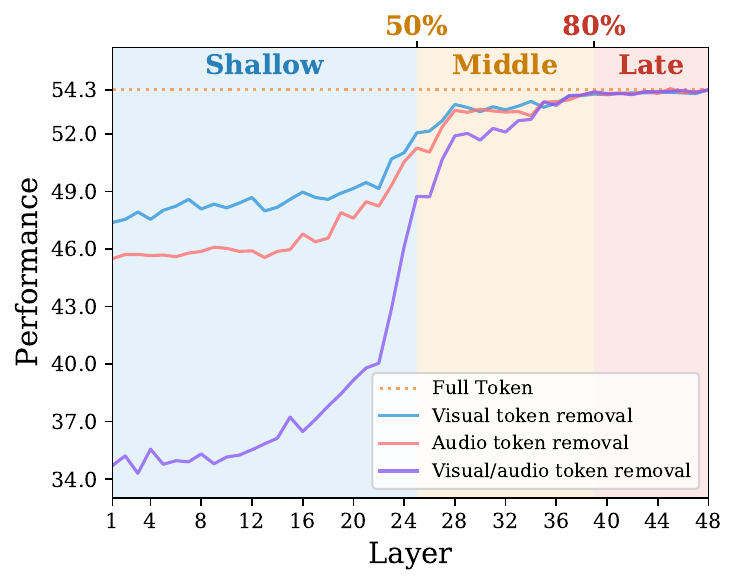}
        \caption{Qwen3-Omni-30B}
        \label{fig:early_remove_performance_qw3o}
    \end{subfigure}

    \caption{
    \textbf{The impact of full visual / audio token removal on the performance of an om-LLM}. Test set: WorldSense.  Depending on the impact, we roughly divide the LLM layers into three blocks: \emph{shallow} layers, where removal causes a collapse in model performance, \emph{middle} layers, where removal leads to moderate performance loss, and \emph{late} layers, where removal has no impact on performance.
    }
    \label{fig:early_remove_worldsense}
    \vspace{-10pt}
\end{figure}

%% file: sec/4_method.tex
\input{figure/method_overview}

As illustrated in~\cref{fig:method_overview}, \ourmethod is a three-stage method. The first stage performs pre-LLM token selection (\cref{ssec:method_prellm}), the second stage performs inner-LLM token selection (\cref{ssec:method_innerllm}), whilst the last stage simply removes all \emph{non-text} tokens at the late LLM layers.

%subsec+++++++++++++
\subsection{Stage I: Pre-LLM Token Selection by Window-based DivPrune}
\label{ssec:method_prellm}
%subsec+++++++++++++

Much redundancy exists in both visual and audio tokens in the pre-LLM stage. For instance, visual tokens within a given window typically show high inter-token affinity, especially in low-motion regions. In order to select a compact yet diverse subset, we extend DivPrune \citep{alvar2025divprune}, originally proposed for image token selection, to the omni-modal context.
DivPrune selects tokens by greedily solving a max-min diversity problem, where the objective is to maximize the minimum inter-token distance within the selected subset. To that end, a token-wise distance matrix is computed. We adapt DivPrune for omni-modal token selection as follows. First, for efficiency, instead of computing the distance matrix for all input tokens, we restrict the computation to a per-window and per-modality basis. Second, to encourage the selection of salient tokens, the matrix is row-wise reweighed by each token's attention scores. We term the adapted DivPrune \prellm.

Recall that in our design, the \trr progressively goes down as the tokens propagate forward. Therefore, the pre-LLM \trr, denoted by $r_s$, shall be larger than $R$. To this end, letting $r_{s,v}$ and $r_{s,a}$ be the visual and audio pre-LLM \trrs, respectively, we set  $r_{s,v}=\lambda R_v$ and $r_{s,a}=\lambda R_a$, where $\lambda>1$ is a pre-specified scale factor. Consequently, after the \prellm operation, the number of non-textual tokens to be forwarded to the LLM is reduced from $N_o$ to $r_{s,v} N_v + r_{s,a} N_a$.

%subsec+++++++++++++
\subsection{Stage II: Inner-LLM Token Selection with Top-down Token Budget Allocation}
\label{ssec:method_innerllm}
%subsec+++++++++++++

%subsubsec=============
\subsubsection{Block-wise \trr Decay Schedule}
\label{ssec:method_exp_schedule}
%subsubsec=============

Based on the pattern of block-wise layer importance (\cref{ssec:analysis_early_remove}),  we roughly divide the $\mathrm{L}$ layers of the LLM into three blocks, \ie \emph{shallow}, \emph{middle}, and \emph{late}, with two hyperparameters $L_{s}$ and $L_l$ indicating the shallow-middle and middle-late boundary layers, respectively. Consequently, we propose a block-wise decay schedule for per LLM-layer \trr allocation, as detailed in~\cref{fig:block_decay_schedule}.

Since layers in the shallow block are critical for cross-modal fusion, no token pruning is performed in these layers. The visual and audio \trrs are kept identical to their pre-LLM counterparts, $r_{s,v}$ and $r_{s,a}$. For notational simplicity, we omit the modality subscript and simply write $r_s$ in the following.

The middle block is responsible for token selection with progressively decayed \trrs. As the layer importance diminishes with depth, deeper layers can afford more aggressive token pruning. For fine-grained \trr allocation, we define alongside $L_s$ two extra \trr-transition layers, $L_{m_1}$ and $L_{m_2}$. Accordingly, the middle block is divided into three sub-blocks with layer ranges $(L_s, L_{m_1}]$, $(L_{m_1}, L_{m_2}]$, and $(L_{m_2}, L_l]$. The \trr decreases across sub-blocks with a linearly increasing step. In particular, let $r_{m_i}$ be the \trr of sub-block $i$ (=1, 2, 3). Our decay schedule is defined as $r_{m_{i}}=r_{m_{i-1}} - i\cdot \delta$, with $r_{m_0}=r_s$ and $\delta$ a scale factor. This schedule enables earlier sub-blocks to undergo relatively mild token pruning while later sub-blocks discard tokens more aggressively, see~\cref{fig:block_decay_schedule}. Note that with $\lambda$ provided, $\delta$ can be computed analytically (Appendix~\ref{app:method_details}).
All the transition layers, \ie $\{L_s, L_{m_1}, L_{m_2}, L_l\}$, are selected using a fixed layer-to-$\mathrm{L}$ ratio, \ie $\lfloor \mathrm{L}\cdot p \rfloor + 1$, where $p$ is taken from $\{0.5,0.6,0.7,0.8\}$, respectively (\cref{table:transition_layer}).

\input{figure/fig_block_decay_schedule}

%subsubsec=============
\subsubsection{Top-down Token Budget Allocation}
\label{ssec:method_mmwin}
%subsubsec=============

For each layer in the middle block, substituting $R_v$ and $R_a$ for $R$ in~\cref{fig:block_decay_schedule} yields its visual and audio \trrs, denoted as $r_v$  and $r_a$, respectively. The layer then accepts $r_v N_v$ visual tokens and $r_a N_a$ audio tokens as input. 
Recall that the input tokens are grouped into windows along the temporal dimension. Intuitively, windows containing more relevant information \wrt to the user query should be allocated a higher token budget. Similarly, within every window, the modality (visual or audio) that is more relevant \wrt the user query should also receive a larger budget relative to the other modality. In that regard, we propose a top-down strategy for query-guided token budget allocation.

\textbf{Inter-window token budget allocation}. For each window $t$ (=$1,\ldots,T$), we measure its relevance to the user query, denoted as $S_t$, based on the causal self-attention scores between the query and the visual and audio tokens within the window. Specifically, the query is represented by the last textual token, which has attended to all preceding tokens under causal attention. The visual-based window-query relevance score $S_{t,v}$ is computed as the mean of the query-to-visual-tokens attention scores, and then normalized using a softmax.
In a similar manner, we obtain the audio-based relevance score $S_{t,a}$. The overall window-query relevance $S_t$ is then defined as the average of $S_{t,v}$ and $S_{t,a}$. The  token budget $B_t$ allocated to window $t$ is computed as $(r_v N_v + r_a N_a) S_t$.

\textbf{Intra-window token budget re-allocation}.
For token budget re-allocation within each window, we jointly consider each modality's layer-wise budget and its relevance to the query, computing the window-wise visual and audio token budgets, $B_{t,v}$ and $B_{t,a}$, as follows:
\begin{equation}\label{eq:intra_window_score_split}
\left\{
    \begin{array}{ll}
        B_t & = (r_v \cdot  N_v + r_a \cdot  N_a) S_t \\ 
        B_{t,v} & = \dfrac{S_{t,v}\cdot r_v \cdot  N_v}{S_{t,v}\cdot r_v \cdot  N_v + S_{t,a}\cdot r_a \cdot  N_a} B_t \\ 
        B_{t,a} & = B_t - B_{t,v}.
    \end{array}
    \right.
\end{equation}
Note that if~\cref{eq:intra_window_score_split} does not fully allocate the budget, the remaining tokens will be re-allocated proportionally to $S_t$ to 
ensure $\sum_{t=1}^T B_t = r_v N_v + r_a N_a$.

%subsubsec=============
\subsubsection{Query-guided Visual and Audio Token Selection}
\label{ssec:method_text_guided_selection}
%subsubsec=============

In order to select $B_{t,v}$ visual tokens from window $t$, we sort the visual tokens in descending order by the previously computed query-to-visual-tokens attention scores, and consequently retain the top $B_{t,v}$ tokens. Audio tokens are selected in a similar vein.

%% file: figure/method_overview.tex
\begin{figure*}[t]
    \centering
    \includegraphics[width=1.0\textwidth]{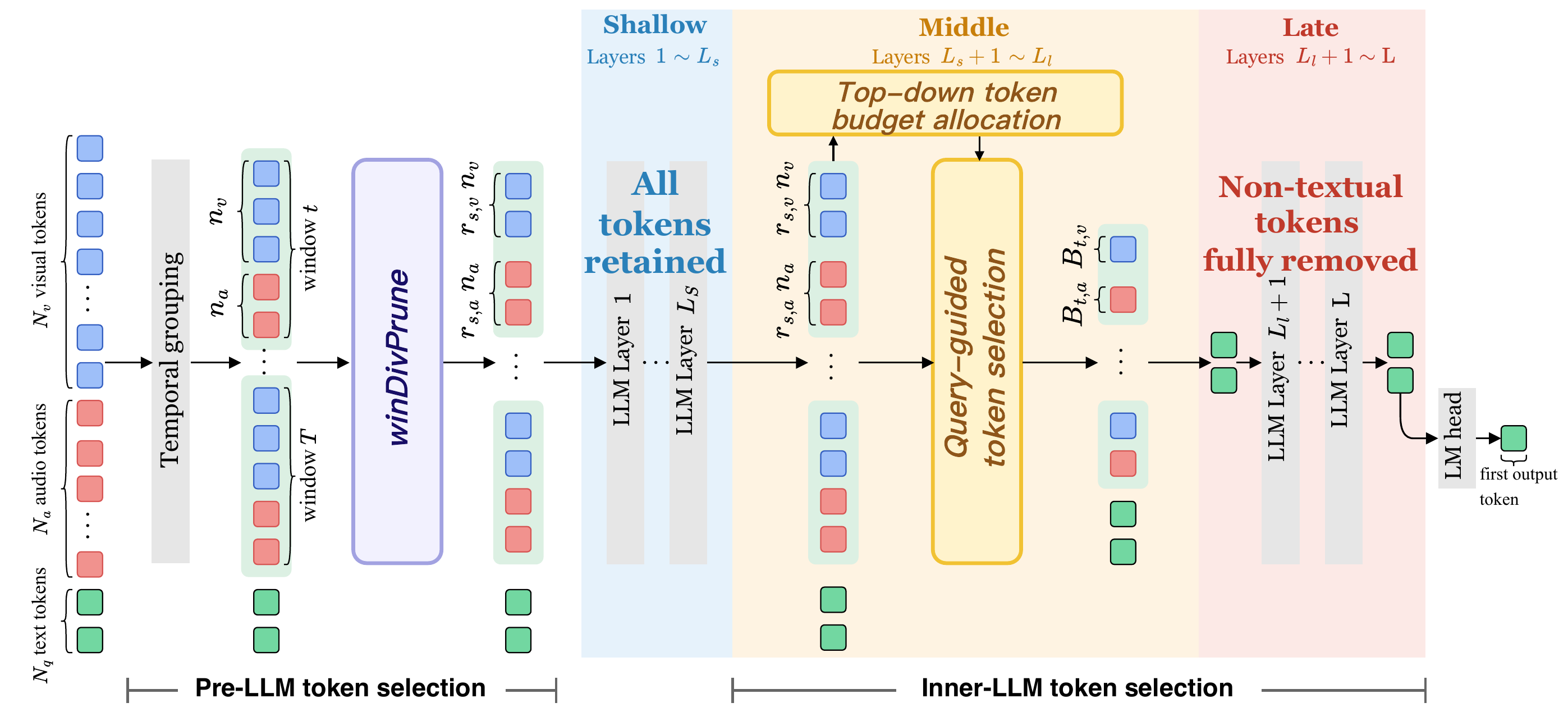}
    \caption{
    \textbf{\ourmethod for om-LLM token selection}. Given an $\mathrm{L}$-layer om-LLM,  the boundaries between the shallow, middle and late blocks are determined in terms of prespecified ratios (derived from \cref{fig:early_remove_worldsense}), \ie $L_s=0.5\times \mathrm{L}$ and $L_l=0.8\times \mathrm{L}$. Such a simple block partition works for Qwen2.5-Omni-7B ($\mathrm{L}=28$), OmniVinci-9B ($\mathrm{L}=28$) and Qwen3-Omni-30B ($\mathrm{L}=48$). 
    }
    \label{fig:method_overview}
    \vspace{-10pt}
\end{figure*}

%% file: figure/fig_block_decay_schedule.tex
\definecolor{shallowBG}{HTML}{3498db}
\definecolor{middleBG}{HTML}{f39c12}
\definecolor{lateBG}{HTML}{e74c3c}
\definecolor{shallowTXT}{HTML}{2980b9}
\definecolor{middleTXT}{HTML}{c87f0a}
\definecolor{lateTXT}{HTML}{c0392b}
\newcommand{\srowc}{\cellcolor{shallowBG!12}}
\newcommand{\mrowc}{\cellcolor{middleBG!10}}
\newcommand{\lrowc}{\cellcolor{lateBG!12}}
\begin{figure}[htbp!]
\centering
\begin{minipage}[t]{0.46\linewidth}
    \vspace{0pt}
    \centering
    \includegraphics[width=\linewidth]{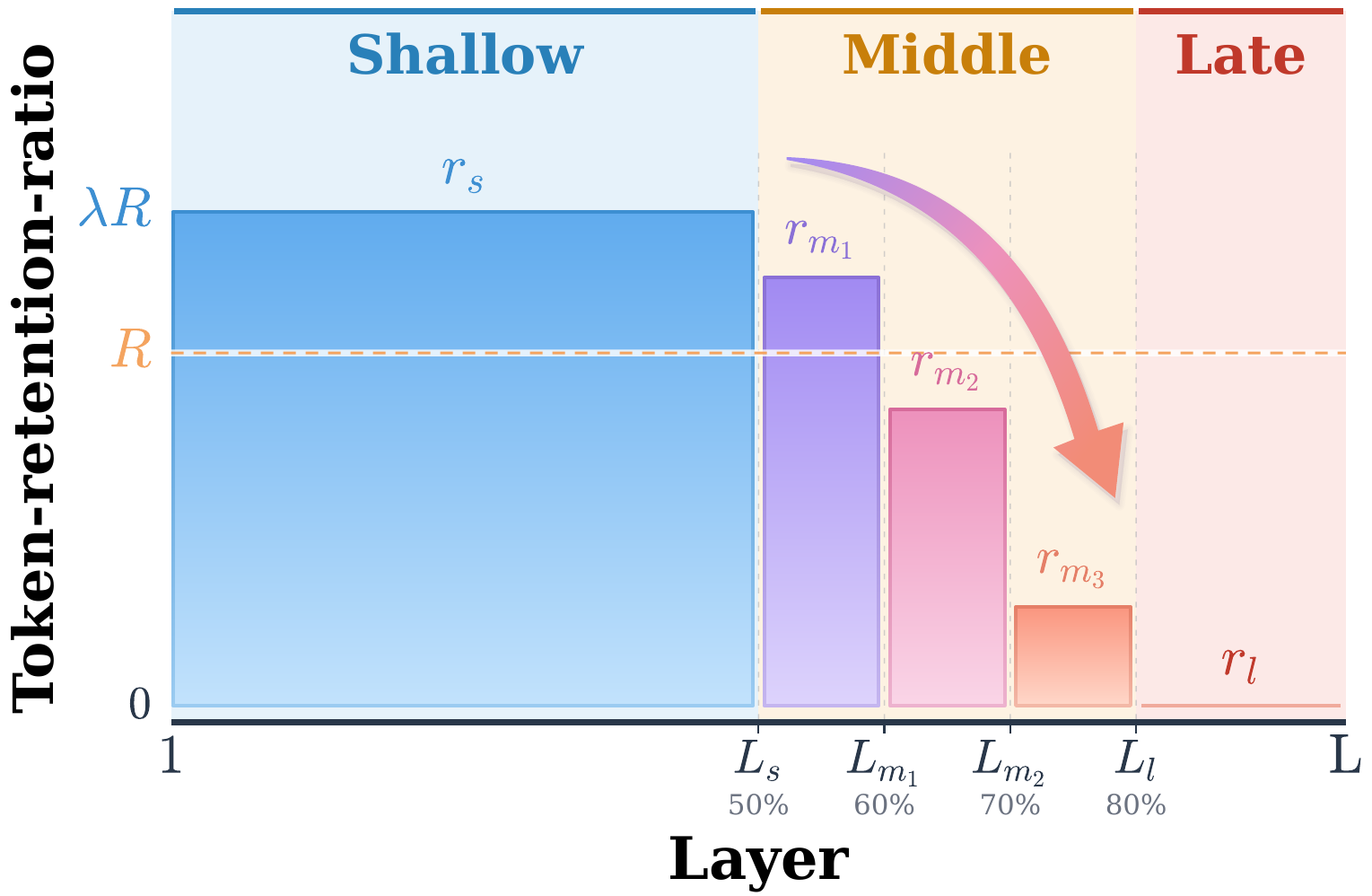}
\end{minipage}
\hfill
\begin{minipage}[t]{0.52\linewidth}
    \vspace{1pt}  % 用于top line对齐
    \centering
    \setlength{\tabcolsep}{0.5pt} % 设置表格中列间距
    \renewcommand{\arraystretch}{1.21} % 调整表格行高度
    \resizebox{\linewidth}{!}{%
    \begin{tabular}{l l l l l l}
    \toprule

    \textbf{Block} & &
    \textbf{Layer range} & &
    \multicolumn{2}{l}{\textbf{\trr per LLM-layer}} \\ [1pt]
    % \midrule
    \specialrule{0.8pt}{0pt}{0pt}  % 实线黑色, 加粗（连续，避免 hdashline 逐列断裂）

    \srowc \textcolor{shallowTXT}{Shallow} & \srowc\hspace{10pt} & \srowc $[1,\, L_s]$ & \srowc\hspace{10pt} & \srowc $r_s$ & \srowc ${}=\lambda \cdot R$ \\ [1pt]
    \noalign{\hbox{\color{gray}\hdashrule{6.9cm}{0.4pt}{2pt 2pt}}}  % dashrule 虚线：段长 2pt / 间隔 2pt

    \mrowc & \mrowc\hspace{10pt} & \mrowc $(L_{s},\, L_{m_1}]$ & \mrowc\hspace{10pt} & \mrowc $r_{m_1}$ & \mrowc ${}=r_s-\delta$ \\
    \mrowc \textcolor{middleTXT}{Middle} & \mrowc\hspace{10pt} & \mrowc $(L_{m_1},\, L_{m_2}]$ & \mrowc\hspace{10pt} & \mrowc $r_{m_2}$ & \mrowc ${}=r_{m_1} - 2\delta$ \\
    \mrowc & \mrowc\hspace{10pt} & \mrowc $(L_{m_2},\, L_l]$ & \mrowc\hspace{10pt} & \mrowc $r_{m_3}$ & \mrowc ${}=r_{m_2} - 3\delta$ \\ [2pt]
    \noalign{\hbox{\color{gray}\hdashrule{6.9cm}{0.4pt}{2pt 2pt}}}  % dashrule 虚线：段长 2pt / 间隔 2pt

    \lrowc \textcolor{lateTXT}{Late} & \lrowc\hspace{10pt} & \lrowc $(L_l,\, \mathrm{L}]$ & \lrowc\hspace{10pt} & \lrowc $r_l$ & \lrowc ${}=0$ \\ [1.5pt]
    \specialrule{0.8pt}{0pt}{0pt}  % 实线黑色, 加粗（连续，避免 hdashline 逐列断裂）
    \end{tabular}
    }
\end{minipage}
\caption{\textbf{Proposed block-wise decay schedule for per-LLM-layer \trr}.}
\label{fig:block_decay_schedule}
\vspace{-10pt}
\end{figure}

%% file: sec/5_experiment.tex
%subsec+++++++++++++
\subsection{Experimental Setup}
\label{ssec:experimental_setup}
%subsec+++++++++++++

\textbf{Test sets}.
We evaluate \ourmethod on the following six test sets, commonly used to evaluate an om-LLM's audio-visual understanding abilities:
\textit{WorldSense}~\citep{hong2025worldsense}, \textit{Daily-Omni}~\citep{zhou2025dailyomni}, \textit{AVUT}~\citep{Yang2025avut}, \textit{OmniVideoBench}~\citep{li2025omnivideobench}, \textit{Video-MME}~\citep{fu2025videomme}, and \textit{LVOmniBench}~\citep{tao2026lvomnibench}.

\textbf{Choice of om-LLM}. We experiment with three open-source om-LLMs,
\ie Qwen2.5-Omni-7B (28-layer LLM) \citep{xu2025qwen25omni}, OmniVinci-9B (28-layer LLM) \citep{ye2025omnivinci}, and Qwen3-Omni-30B (A3B-Instruct, 48-layer MoE-based LLM) \citep{xu2025qwen3omni}. Note that Qwen3-Omni-30B has an audio token rate of 13 tokens per second, lower than Qwen2.5-Omni-7B's and OmniVinci's 25 tokens per second. Consequently, for the same overall \trr ($R$), the visual \trr ($R_v$) and audio \trr ($R_a$) differ between the om-LLMs.

\input{table/table_qw25o_main_all.tex}

\textbf{Baselines}.
To ensure a fair and reproducible comparison, we opt for baseline methods that are training-free, applicable either before or during the prefill stage, requiring no \textit{external} model, and \textit{open-source}. Eventually, we compile a list of seven recent methods, adapting them as needed for om-LLM. 
Depending on their targeted modalities, \ie image, video or omni-modal, the baselines are categorized into the following three groups: \\
$\bullet$ \emph{Image}: FastV \citep{chen2024fastv}, VisionZip \citep{yang2025visionzip} and DivPrune\citep{alvar2025divprune}. Applying each method in parallel to visual and audio tokens yields an omni-modal variant that we refer to as FastV-om, VisionZip-om, and DivPrune-om, respectively. \\
$\bullet$ \emph{Video}:  DyCoke~\citep{tao2025dycoke}, FastVID~\citep{shenfastvid}. Following \citep{tao2025omnizip,ding2026omnisift}, for DyCoke we use its prefill-stage TTM module only. \\
$\bullet$ \emph{Omni-modal}: OmniZip~\citep{tao2025omnizip}, DASH~\citep{li2025dash} and Random that randomly selects tokens at a given ratio.

\textbf{Implementation}. We refer to Appendix~\ref{app:method_details}.

%subsec+++++++++++++
\subsection{\ourmethod \emph{versus} SOTA}
\label{ssec:main_results}
%subsec+++++++++++++

\textbf{Results on Qwen2.5-Omni-7B}.
As shown in~\cref{table:qw25o_main_all}, \ourmethod achieves the best average performance across all retention ratios, surpassing the full-token baseline at 35\% (51.4 \vs 51.2) and retaining 94.9\% of it at 10\% with only 11\% of the FLOPs.
Among the baselines, FastV and FastV-om rank lowest, indicating that shallow-layer relevance scores are too noisy for reliable one-shot pruning, whereas VisionZip-om and DivPrune-om match OmniZip and DASH, suggesting that pre-LLM local saliency selection already captures information comparable to joint audio-visual compression.
Daily-Omni is the exception, as its reliance on audio evidence allows video-only methods that keep all audio tokens to score higher, yet \ourmethod still reaches the best mean score.

\input{table/table_efficiency}

\textbf{Efficiency analysis} is provided in~\cref{table:efficiency}.
The reported prefill time includes inner-LLM token selection overhead, whereas TTFT further accounts for encoder forward and pre-LLM compression.
Owing to careful code optimization with vectorized tensor operations, the token selection overhead of \ourmethod remains well controlled across retention ratios.
At 35\% retention, \ourmethod achieves a $2.2\times$ prefill speedup ($1.5\times$ in TTFT), while attaining the highest mean score (51.4 \vs 51.2 for full tokens).
At 10\% retention, the prefill speedup further increases to $4.9\times$ ($1.9\times$ in TTFT) with GPU peak memory lowered to 17.68\,GB, demonstrating a favorable efficiency-performance trade-off.

\textbf{Results on Qwen3-Omni-30B}.
Qwen3-Omni-30B employs an MoE architecture with a deeper LLM and a DeepStack mechanism that injects multi-scale visual features into the early LLM layers. Accordingly, \ourmethod synchronizes the pruning of these injected features. 
As reported in~\cref{table:qw3o_main_all}, our method achieves the best average performance across all retention ratios, preserving 99.8\% of the full-token performance at 35\% retention and 94.7\% at the more aggressive 10\% \trr. The results confirm the effectiveness of \ourmethod on a more complex MoE architecture.

\textbf{Results on OmniVinci-9B}.
As shown in~\cref{table:ovinci_main_all}, \ourmethod consistently achieves the best average performance under different \trrs, indicating strong generalization across diverse om-LLMs.

%subsec+++++++++++++
\subsection{Ablation Studies}
\label{ssec:ablation_studies}
%subsec+++++++++++++

Ablation studies are conducted on Qwen2.5-Omni with $R{=}35\%$. See~\cref{table:ablation_overall} and~\cref{fig:param_ablation,fig:boundary_sensitivity}.

\input{table/ablation/table_ablation_overall}
\input{figure/fig_ablation_sensitivity+lambda}

\textbf{\texttt{WinDivPrune} for pre-LLM token selection.}
Removing either attention weighting or diversity degrades performance (Setup-2 and Setup-3), confirming that the two are complementary. 
Replacing \prellm with random selection (Setup-1) or OmniZip (Setup-4) leads to a larger drop.

\textbf{\trr decay schedule.}
Replacing our linear-step block-wise decay schedule with one-shot selection at a single layer (Setup-5) or an equal-step schedule (Setup-6) results in performance loss.

\textbf{Top-down allocation for token budget.}
Replacing per-window selection with global ranking (Setup-7) lowers performance, confirming window-local selection better preserves temporal structure. Decoupling modality budgets (Setup-8), removing inter-window (Setup-9), or intra-window re-allocation (Setup-10) each degrades performance, verifying every level contributes.

\textbf{Choice of $\lambda$.}
\cref{fig:param_ablation} shows that increasing $\lambda$ generally improves performance, indicating that allocating more tokens to relevance-based inner-LLM compression is more effective than pre-LLM pruning.

\textbf{Robustness to block boundaries.}
\cref{fig:boundary_sensitivity} reveals that shifting the transition block boundaries by $\pm 2$ layers has minimal impact on the mean score, confirming robust boundary choices.

%% file: table/table_qw25o_main_all.tex
\begin{table}[t!]
\caption{\textbf{Results on Qwen2.5-Omni-7B}. Per method, the visual token retention ratio $R_v$ and the audio counterpart $R_a$ are adjusted to satisfy the given overall ratio $R$. \textbf{Bold} and \underline{underline} denote the best and second-best per column. Methods sorted in ascending order by their mean performance. }
\label{table:qw25o_main_all}

\centering
\setlength{\tabcolsep}{3.5pt} % 设置表格中列间距
\renewcommand{\arraystretch}{1.0} % 调整表格行高度

\resizebox{1.0\linewidth}{!}{
\begin{tabular}{@{}l| l c | cccccc c@{}}
\toprule
\textbf{Method} & $R$ (\textcolor{vcolor}{$R_v$}-\textcolor{acolor}{$R_a$}) & \textbf{TFLOPs$\downarrow$} & \textbf{\specialcell{World\\Sense}} & \textbf{\specialcell{Daily\\Omni}} & \textbf{\specialcell{AVUT}} & \textbf{\specialcell{OmniVideo\\Bench}} & \textbf{\specialcell{Video\\MME}} & \textbf{\specialcell{LVOmni\\Bench}} & \textbf{Mean} \\
\midrule

Full tokens &
100 \var{100,100} & 111.0 &
46.7 & 64.0 & 63.6 & 34.1 & 65.3 & 33.3
& 51.2$_{\text{\;\scriptsize 100.0\%}}$ \\

\midrule

%%%%%%%%%%%%%%  35 Group  %%%%%%%%%%%%%%
FastV-om &
35 \var{30,65} & 38.3 &
43.3 & 58.7 & 59.2 & 34.4 & 65.6 & 34.3
& 49.3$_{\text{\;\scriptsize 96.3\%}}$ \\

FastV \tiny{ECCV'24} &
35 \var{24,100} & 38.3 &
43.7 & 59.6 & 59.1 & 34.7 & 65.2 & 34.8
& 49.5$_{\text{\;\scriptsize 96.7\%}}$ \\

Random &
35 \var{30,65} & 37.4 &
44.6 & 59.7 & 61.1 & 34.1 & 65.0 & 34.9
& 49.9$_{\text{\;\scriptsize 97.5\%}}$ \\

DyCoke \tiny{CVPR'25} &
35 \var{24,100} & 37.7 &
44.4 & 59.9 & 60.3 & 34.6 & 66.0 & 34.7
& 50.0$_{\text{\;\scriptsize 97.7\%}}$ \\

FastVID \tiny{NIPS'25} &
35 \var{24,100} & 37.1 &
44.6 & 59.8 & 60.4 & 34.6 & 65.4 & \underline{35.8}
& 50.1$_{\text{\;\scriptsize 97.9\%}}$ \\

VisionZip \tiny{CVPR'25} &
35 \var{24,100} & 37.4 &
44.5 & 60.6 & 60.7 & 33.6 & 66.2 & 35.3
& 50.2$_{\text{\;\scriptsize 98.0\%}}$ \\

DivPrune \tiny{CVPR'25} &
35 \var{24,100} & 37.4 &
44.8 & \underline{60.7} & 61.0 & 33.9 & 66.1 & 35.7
& 50.4$_{\text{\;\scriptsize 98.4\%}}$ \\

OmniZip \tiny{CVPR'26} &
35 \var{30,65} & 38.2 &
45.2 & 60.4 & 61.3 & 34.5 & 66.3 & 34.7
& 50.4$_{\text{\;\scriptsize 98.4\%}}$ \\

DASH \tiny{ECCV'26} &
35 \var{30,65} & 37.8 &
44.7 & 59.9 & 61.3 & \underline{34.8} & \underline{66.5} & 35.4
& 50.4$_{\text{\;\scriptsize 98.4\%}}$ \\

DivPrune-om &
35 \var{30,65} & 37.4 &
\underline{45.4} & 59.7 & 61.7 & 34.7 & 66.3 & 35.4
& 50.5$_{\text{\;\scriptsize 98.6\%}}$ \\

VisionZip-om &
35 \var{30,65} & 37.4 &
45.2 & 60.2 & \underline{62.3} & 34.6 & 66.3 & 35.4
& \underline{50.7}$_{\text{\;\scriptsize 99.0\%}}$ \\

% \rowcolor{green!8}
\rowcolor[HTML]{f0edff}
\ourmethod &
35 \var{30,65} & 36.5 &
\textbf{46.3} & \textbf{62.1} & \textbf{62.7} & \textbf{35.0} & \textbf{66.6} & \textbf{36.0}
& \textbf{51.4}$_{\text{\;\scriptsize 100.4\%}}$ \\

\midrule

%%%%%%%%%%%%%%  25 Group  %%%%%%%%%%%%%%
FastV &
25 \var{12,100} & 28.3 &
42.4 & 56.5 & 54.2 & 33.4 & 62.8 & 33.8
& 47.2$_{\text{\;\scriptsize 92.2\%}}$ \\

FastV-om &
25 \var{20,55} & 28.3 &
41.6 & 56.1 & 57.6 & 33.0 & 63.4 & 32.7
& 47.4$_{\text{\;\scriptsize 92.6\%}}$ \\

\graytxt{OmniZip} &
\graytxt{25 \var{25,25}} & \graytxt{23.9} &
\graytxt{42.7} & \graytxt{55.5} & \graytxt{57.7} & \graytxt{32.2} & \graytxt{66.0} & \graytxt{34.0} &
\graytxt{48.0$_{\text{\;\scriptsize 93.8\%}}$} \\

Random &
25 \var{20,55} & 27.6 &
43.0 & 57.3 & 59.1 & 33.5 & 64.2 & 34.6
& 48.6$_{\text{\;\scriptsize 94.9\%}}$ \\

FastVID &
25 \var{12,100} & 27.0 &
43.4 & 58.5 & 57.8 & 33.8 & 64.3 & 34.8
& 48.8$_{\text{\;\scriptsize 95.3\%}}$ \\

VisionZip &
25 \var{12,100} & 27.6 &
42.4 & 59.7 & 58.4 & 34.4 & 64.1 & 34.5
& 48.9$_{\text{\;\scriptsize 95.5\%}}$ \\

VisionZip-om &
25 \var{20,55} & 27.6 &
44.4 & 58.3 & 59.1 & 33.7 & 65.4 & 34.7
& 49.3$_{\text{\;\scriptsize 96.3\%}}$ \\

DivPrune &
25 \var{12,100} & 27.6 &
43.2 & \underline{59.7} & 58.4 & 34.3 & 65.3 & \underline{35.9}
& 49.5$_{\text{\;\scriptsize 96.7\%}}$ \\

DASH &
25 \var{20,55} & 27.6 &
44.3 & 58.7 & 59.6 & 34.3 & 66.2 & 35.5
& 49.8$_{\text{\;\scriptsize 97.3\%}}$ \\

DivPrune-om &
25 \var{20,55} & 27.6 &
\underline{44.5} & 59.2 & \underline{60.4} & \textbf{34.5} & \underline{66.2} & 35.4
& \underline{50.0}$_{\text{\;\scriptsize 97.7\%}}$ \\

\rowcolor[HTML]{f0edff}
\ourmethod &
25 \var{20,55} & 26.4 &
\textbf{45.1} & \textbf{60.9} & \textbf{62.5} & \underline{34.4} & \textbf{66.3} & \textbf{36.0}
& \textbf{50.9}$_{\text{\;\scriptsize 99.4\%}}$ \\

\midrule

%%%%%%%%%%%%%%  15 Group  %%%%%%%%%%%%%%
FastV-om &
15 \var{10,45} & 24.3 &
38.4 & 51.8 & 51.0 & 31.2 & 58.8 & 31.2
& 43.7$_{\text{\;\scriptsize 85.4\%}}$ \\

Random &
15 \var{10,45} & 18.0 &
40.4 & 55.7 & 57.2 & 32.7 & 63.4 & 32.8
& 47.0$_{\text{\;\scriptsize 91.8\%}}$ \\

VisionZip-om &
15 \var{10,45} & 17.6 &
43.0 & 57.1 & 56.5 & 33.6 & 63.1 & 33.8
& 47.9$_{\text{\;\scriptsize 93.6\%}}$ \\

DASH &
15 \var{10,45} & 18.0 &
\underline{42.9} & 56.2 & 57.3 & \underline{33.7} & 64.2 & 34.1
& 48.1$_{\text{\;\scriptsize 93.9\%}}$ \\

DivPrune-om &
15 \var{10,45} & 17.6 &
42.4 & \underline{57.6} & 57.7 & 33.0 & \underline{64.4} & \textbf{34.4}
& \underline{48.3}$_{\text{\;\scriptsize 94.3\%}}$ \\

\rowcolor[HTML]{f0edff}
\ourmethod &
15 \var{10,45} & 17.3 &
\textbf{43.9} & \textbf{58.7} & \textbf{60.2} & \textbf{34.2} & \textbf{65.9} & \underline{34.2}
& \textbf{49.5}$_{\text{\;\scriptsize 96.7\%}}$ \\

\midrule

%%%%%%%%%%%%%%  10 Group  %%%%%%%%%%%%%%
Random &
10 \var{6,35} & 12.8 &
39.2 & 52.0 & 54.0 & 31.6 & 60.4 & 32.9
& 45.0$_{\text{\;\scriptsize 87.9\%}}$ \\

VisionZip-om &
10 \var{6,35} & 12.8 &
\underline{41.8} & 52.8 & 53.8 & 33.4 & 60.7 & 34.4
& 46.2$_{\text{\;\scriptsize 90.2\%}}$ \\

DivPrune-om &
10 \var{6,35} & 12.8 &
41.3 & \underline{55.1} & \underline{55.9} & \textbf{33.5} & \underline{63.4} & \underline{34.8}
& \underline{47.3}$_{\text{\;\scriptsize 92.4\%}}$ \\

\rowcolor[HTML]{f0edff}
\ourmethod &
10 \var{6,35} & 12.0 &
\textbf{43.4} & \textbf{57.2} & \textbf{58.5} & \underline{33.4} & \textbf{63.8} & \textbf{35.4}
& \textbf{48.6}$_{\text{\;\scriptsize 94.9\%}}$ \\

\bottomrule
\end{tabular}
}
\vspace{-5pt}
\end{table}

%% file: table/table_efficiency.tex
\begin{table}[t!]
\caption{
\textbf{Efficiency analysis on Qwen2.5-Omni-7B}.
GPU usage, the time spans for token selection and prefill, and time-to-first-token (TTFT) are all measured on WorldSense using an A800 GPU.
}
\label{table:efficiency}
\centering
\setlength{\tabcolsep}{4.0pt} % 设置表格中列间距
\renewcommand{\arraystretch}{1.05} % 调整表格行高度

\resizebox{0.92\linewidth}{!}{
\begin{tabular}{@{}lrrc c rr c ccl@{}}
\toprule

\multirow{2}{*}{\textbf{Method}} &
\multirow{2}{*}{$R$} &
\multicolumn{1}{c}{\multirow{2}{*}{\specialcell{\textbf{TFLOPs}$\downarrow$}}} &
\multirow{2}{*}{\specialcell{\textbf{GPU Peak}$\downarrow$ \\ \textbf{Mem.} (GB)}} &
&
\multicolumn{2}{c}{\textbf{Token selection} (sec.)$\downarrow$} & &
\multirow{2}{*}{\textbf{Prefill} (sec.)$\downarrow$} &
\multirow{2}{*}{\textbf{TTFT} (sec.)$\downarrow$} &
\multicolumn{1}{c}{\multirow{2}{*}{\specialcell{\textbf{Mean}$\uparrow$ \\\textbf{Score}}}} \\
\cmidrule{6-7}

& & & & & \textit{Pre-LLM} & \textit{Inner-LLM} & & & & \\
\midrule

\rowcolor{gray!15}
Full tokens & 100
& 111.0~$(1.0\times)$ & 22.83 & & -- & -- & & 0.937~$(1.0\times)$ & 1.471~$(1.0\times)$ &
51.2$_{\text{\;\scriptsize 100.0\%}}$ \\ [2pt]

FastV-om & 35
& 38.3~$(2.9\times)$ & 19.48 & & -- & 0.010 & & 0.363~$(2.6\times)$ & 0.901~$(1.6\times)$ &
49.3$_{\text{\;\scriptsize 96.3\%}}$ \\

VisionZip-om & 35
& 37.4~$(3.0\times)$ & 19.63 & & 0.125 & -- & & 0.349~$(2.7\times)$ & 1.066~$(1.4\times)$ &
50.7$_{\text{\;\scriptsize 99.0\%}}$ \\

DivPrune-om & 35
& 37.4~$(3.0\times)$ & 19.63 & & 0.025 & -- & & 0.349~$(2.7\times)$ & 0.952~$(1.5\times)$ &
50.5$_{\text{\;\scriptsize 98.6\%}}$ \\

OmniZip & 35
& 38.2~$(2.9\times)$ & 19.69 & & 0.095 & -- & & 0.360~$(2.6\times)$ & 1.088~$(1.4\times)$ &
50.4$_{\text{\;\scriptsize 98.4\%}}$ \\

DASH & 35
& 37.8~$(2.9\times)$ & 19.70 & & 0.202 & -- & & 0.350~$(2.7\times)$ & 1.102~$(1.3\times)$ &
50.4$_{\text{\;\scriptsize 98.4\%}}$ \\

\rowcolor[HTML]{f0edff}
\ourmethod & 35
& 36.5~$(3.0\times)$ & 18.70 & & 0.034 & 0.083 & & 0.424~$(2.2\times)$ & 1.011~$(1.5\times)$ &
\textbf{51.4}$_{\text{\;\scriptsize 100.4\%}}$ \\

\rowcolor[HTML]{f0edff}
\ourmethod & 25
& 26.4~$(4.2\times)$ & 18.32 & & 0.028 & 0.076 & & 0.333~$(2.8\times)$ & 0.916~$(1.6\times)$ &
50.9$_{\text{\;\scriptsize 99.4\%}}$ \\

\rowcolor[HTML]{f0edff}
\ourmethod & 10
& 12.0~$(\textbf{9.3}\times)$ & \textbf{17.68} & & 0.020 & 0.059 & & 0.192~$(\textbf{4.9}\times)$ & 0.766~$(\textbf{1.9}\times)$ &
48.6$_{\text{\;\scriptsize 94.9\%}}$ \\

\bottomrule
\end{tabular}
}
\vspace{-10pt}
\end{table}

%% file: table/ablation/table_ablation_overall.tex
% Combined ablation: Pre-LLM selection + Inner-LLM allocation
\begin{table}[t]
\caption{\textbf{Ablation studies}. Om-LLM: Qwen2.5-Omni-7B. Overall \trr ($R$): 35\%.}
%retention ratio.}
\label{table:ablation_overall}

\centering
\setlength{\tabcolsep}{3.5pt} % 设置表格中列间距
\renewcommand{\arraystretch}{1} % 调整表格行高度

\resizebox{1.0\linewidth}{!}{%
\begin{tabular}{c l ccccccc}
\toprule
\textbf{\#} & \textbf{Setup} & \textbf{\specialcell{WorldSense}} & \textbf{\specialcell{Daily-Omni}} & \textbf{\specialcell{AVUT}} & \textbf{\specialcell{OmniVideoBench}} & \textbf{\specialcell{Video-MME}} & \textbf{\specialcell{LVOmniBench}} & \textbf{Mean} \\
\midrule

\rowcolor[HTML]{f0edff}
0 & Full-setup
& \textbf{46.3} & \textbf{62.1} & \textbf{62.7} & \textbf{35.0} & \textbf{66.6} & \textbf{36.0} & \textbf{51.4} \\
[2pt]

\multicolumn{9}{@{}l}{\textit{Stage I: Pre-LLM Token Selection:}} \\
1 & \prellm $\rightarrow$ Random
& 45.5 & 61.4 & 61.9 & 34.4 & 66.4 & 34.8 & 50.7 \\

2 & \prellm $\rightarrow$ DivPrune-om
& 45.9 & 61.6 & 62.7 & 34.8 & 66.2 & 35.3 & 51.1 \\

3 & \prellm $\rightarrow$ VisionZip-om
& 45.3 & 61.8 & 62.6 & 34.3 & 66.3 & 35.3 & 51.0 \\

4 & \prellm $\rightarrow$ OmniZip
& 45.5 & 61.8 & 62.4 & 34.9 & 66.4 & 33.6 & 50.8 \\ [2pt]

\multicolumn{9}{@{}l}{\textit{Stage II: Inner-LLM Token Selection:}} \\

\multicolumn{9}{l}{\textit{\trr Decay Schedule:}} \\
5 & Block-wise Decay $\rightarrow$ Single Decay
& 45.7 & 61.5 & 62.3 & 34.5 & 65.8 & 35.2 & 50.8 \\

6 & Linear Step $\rightarrow$ Equal Step
& 45.8 & 61.7 & 62.4 & 34.8 & 66.4 & 35.6 & 51.1 \\ [1pt]

\multicolumn{9}{l}{\textit{Token Budget Allocation:}} \\
7 & Global Token Selection
& 46.3 & 61.6 & 62.0 & 33.9 & 66.3 & 35.1 & 50.9 \\

8 & Modality Decouple
& 45.4 & 62.1 & 62.3 & 34.1 & 66.3 & 34.9 & 50.9 \\

9 & \wo Inter-window
& 46.2 & 61.5 & 62.4 & 34.3 & 66.4 & 35.4 & 51.0 \\

10 & \wo Intra-window
& 46.1 & 61.7 & 62.3 & 34.5 & 66.2 & 35.8 & 51.1 \\

\bottomrule
\end{tabular}%
}
\end{table}

%% file: figure/fig_ablation_sensitivity+lambda.tex
\begin{figure}[t!]
\centering
\begin{minipage}[t]{0.33\linewidth}
    \centering
    \includegraphics[width=\linewidth]{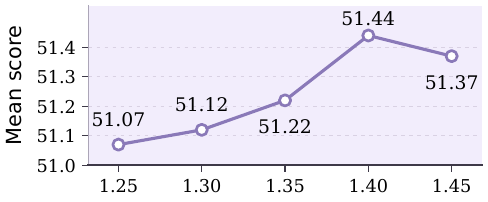}
    \vspace{-15pt}
    \caption{\textbf{Evaluating $\lambda$}.}
    \label{fig:param_ablation}
    \vspace{-10pt}
\end{minipage}
\hfill
\begin{minipage}[t]{0.66\linewidth}
    \centering
    \includegraphics[width=\linewidth]{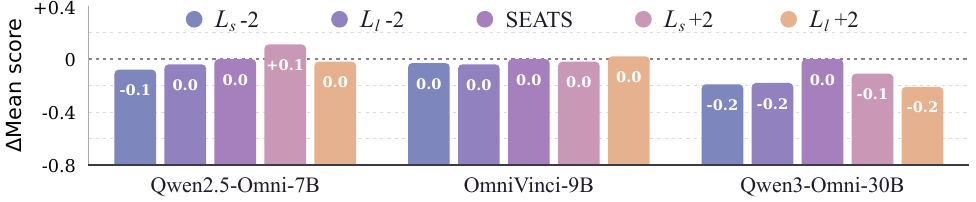}
    \vspace{-15pt}
    \caption{\textbf{Robustness to block boundaries} across three om-LLMs.}
    \label{fig:boundary_sensitivity}
    % \vspace{-10pt}
\end{minipage}
\end{figure}

%% file: sec/6_conclusion.tex
We introduce \ourmethod, a training-free method for efficient om-LLM inference.
Extensive experiments on three omni-modal LLMs (Qwen2.5-Omni-7B, OmniVinci-9B, and Qwen3-Omni-30B) and six audio-visual benchmarks (WorldSense, DailyOmni, AVUT, OmniVideoBench, VideoMME and LVOmniBench) show that \ourmethod achieves a state-of-the-art efficiency-performance trade-off. \ourmethod serves as a plug-and-play module applicable to existing om-LLMs, enabling significant reductions in FLOPs, prefill latency, and memory consumption while preserving task performance.

%% file: sec/7_appendix.tex
\appendix
\input{sec/appendix/0_table_of_contents}
\input{sec/appendix/1_observations}

\input{sec/appendix/4_more_results}
\input{sec/appendix/3_exp_details}
\input{sec/appendix/5_discussion}

%% file: sec/appendix/0_table_of_contents.tex
% Table of Contents for Appendix.
\addtocontents{toc}{\protect\setcounter{tocdepth}{2}}
\setcounter{tocdepth}{-1}
\newpage
\begin{center}
    {\LARGE \textbf{Supplementary Material}}
\end{center}

{\hypersetup{linkcolor=black}% Make links black
\tableofcontents}
\clearpage

%% file: sec/appendix/1_observations.tex
%-----------------------------------------------------------
\section{Generality of the Block-wise Dependency Pattern} \label{sec:app_observations}
%-----------------------------------------------------------

\cref{ssec:analysis_early_remove} reports the block-wise dependency pattern on WorldSense.
To further demonstrate the generality of this pattern across video durations, we present results on Daily-Omni (short videos under one minute,~\cref{fig:early_remove_dailyomni}) and the Video-MME Long subset (videos up to one hour,~\cref{fig:early_remove_videommelong}).
Notably, the block-wise pattern of layer importance holds consistently across all three om-LLMs and datasets, confirming its generality beyond specific benchmarks or model architectures.

\input{figure/early_remove_dailyomni}
\input{figure/early_remove_videommelong}

%% file: figure/early_remove_dailyomni.tex
\begin{figure}[htb]
    \vspace{20pt}
    \centering
    \begin{subfigure}[t]{0.324\linewidth}
        \centering
        \includegraphics[width=\linewidth]{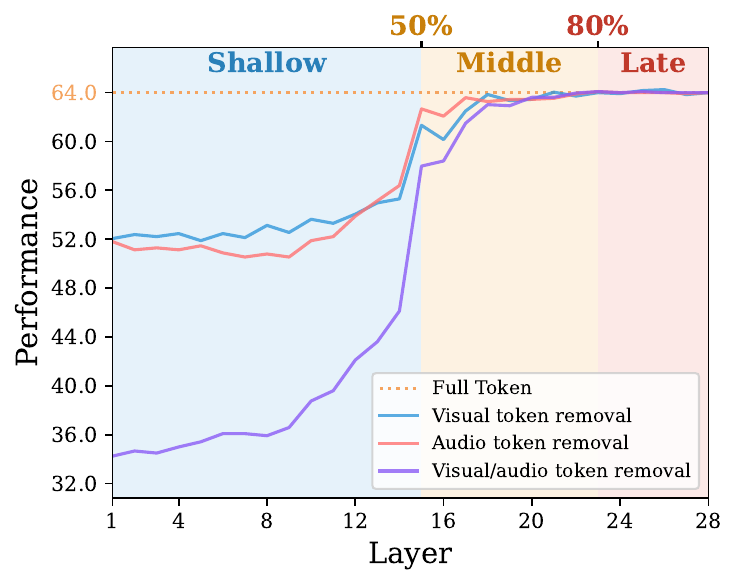}
        \caption{Qwen2.5-Omni-7B}
        \label{fig:early_remove_dailyomni_qw25o}
    \end{subfigure}
    % \hfill
    \hspace{0.001\linewidth}
    \begin{subfigure}[t]{0.324\linewidth}
        \centering
        \includegraphics[width=\linewidth]{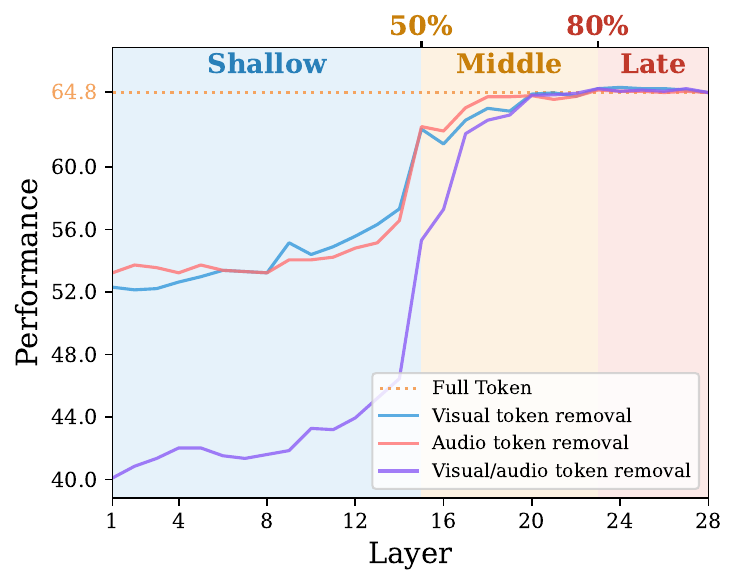}
        \caption{OmniVinci-9B}
        \label{fig:early_remove_dailyomni_omnivinci}
    \end{subfigure}
    % \hfill
    \hspace{0.001\linewidth}
    \begin{subfigure}[t]{0.324\linewidth}
        \centering
        \includegraphics[width=\linewidth]{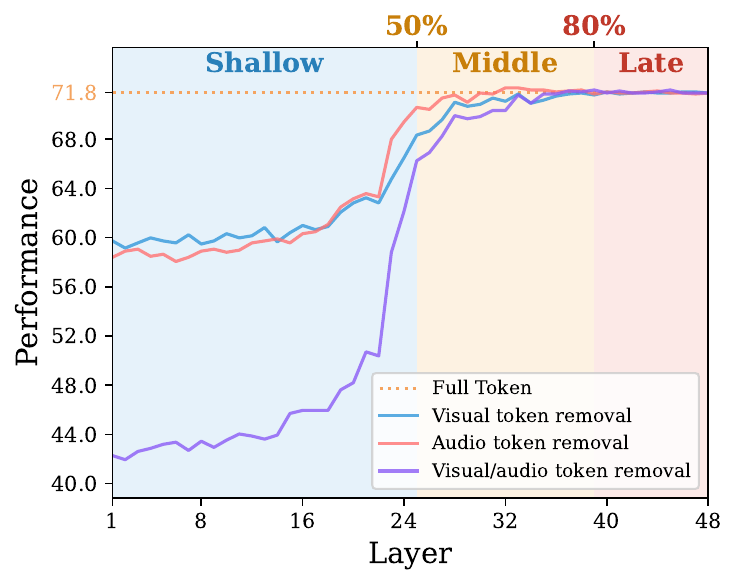}
        \caption{Qwen3-Omni-30B}
        \label{fig:early_remove_dailyomni_qw3o}
    \end{subfigure}

    \caption{
    \textbf{Layer-wise token removal on Daily-Omni, where videos are under one minute}.
    }
    \label{fig:early_remove_dailyomni}
    \vspace{20pt}
\end{figure}

%% file: figure/early_remove_videommelong.tex
\begin{figure}[htb]
    % \vspace{20pt}
    \centering
    \begin{subfigure}[t]{0.324\linewidth}
        \centering
        \includegraphics[width=\linewidth]{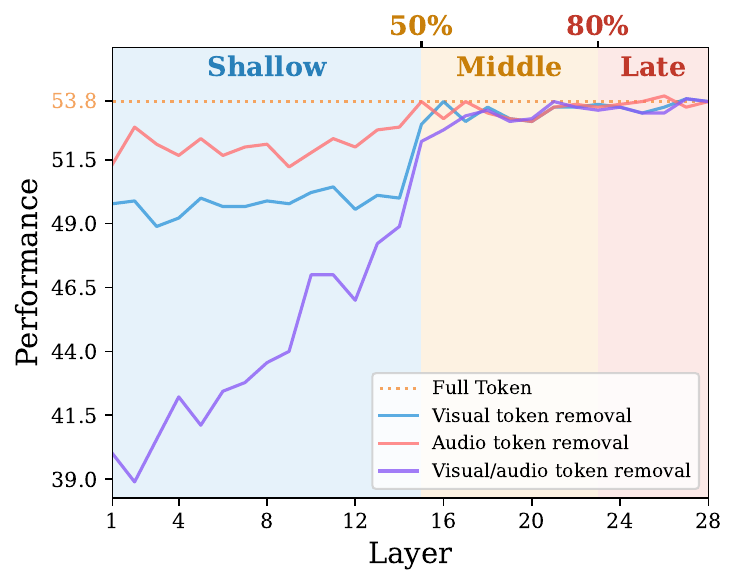}
        \caption{Qwen2.5-Omni-7B}
        \label{fig:early_remove_videommelong_qw25o}
    \end{subfigure}
    % \hfill
    \hspace{0.001\linewidth}
    \begin{subfigure}[t]{0.324\linewidth}
        \centering
        \includegraphics[width=\linewidth]{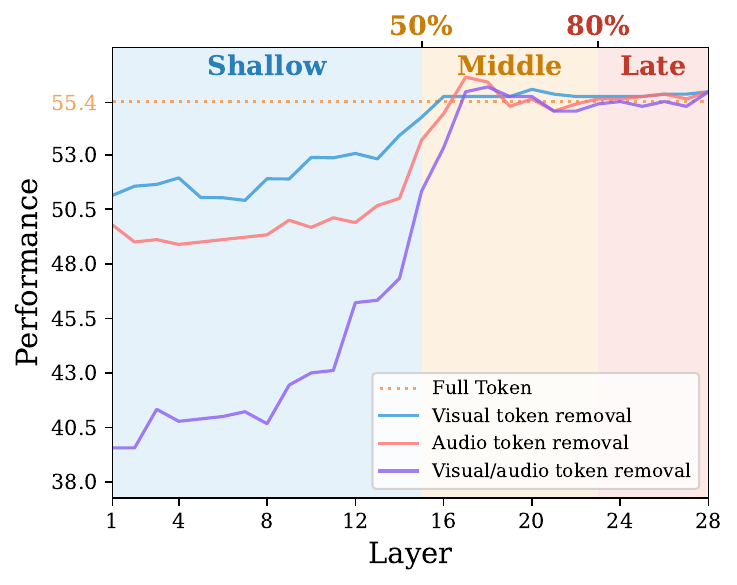}
        \caption{OmniVinci-9B}
        \label{fig:early_remove_videommelong_omnivinci}
    \end{subfigure}
    % \hfill
    \hspace{0.001\linewidth}
    \begin{subfigure}[t]{0.324\linewidth}
        \centering
        \includegraphics[width=\linewidth]{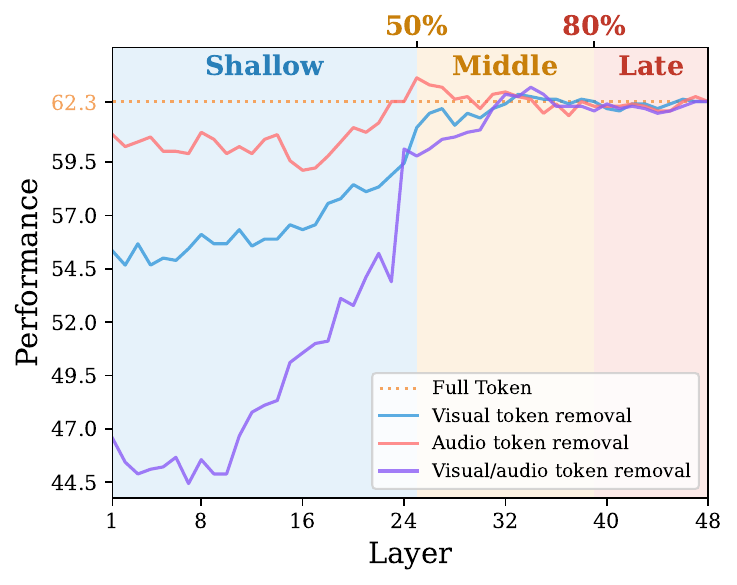}
        \caption{Qwen3-Omni-30B}
        \label{fig:early_remove_videommelong_qw3o}
    \end{subfigure}

    \caption{
    \textbf{Layer-wise token removal on Video-MME Long, where videos are up to one hour}.
    }
    \label{fig:early_remove_videommelong}
    \vspace{20pt}
\end{figure}

%% file: sec/appendix/4_more_results.tex
\clearpage
\input{table/table_qw3o_main_all.tex}

%-----------------------------------------------------------
\section{Additional Experimental Results} \label{sec:app_more_results}
%-----------------------------------------------------------

%subsec+++++++++++++
\subsection{Full Results on Qwen3-Omni-30B} \label{sec:appendix_qw3o_main}
%subsec+++++++++++++
As noted in~\cref{ssec:main_results}, Qwen3-Omni-30B adopts an MoE architecture with a deeper LLM ($\mathrm{L}{=}48$), and \ourmethod synchronizes pruning with its DeepStack feature injection accordingly. \cref{table:qw3o_main_all} presents the full results. The performance gap between \ourmethod and the best baseline widens as $R$ decreases, \ie 99.8\% \vs 98.3\% of the full-token performance at $R{=}35\%$ and 94.7\% \vs 91.3\% at $R{=}10\%$, indicating that our multi-stage design is particularly effective under aggressive compression. The relative ranking of baselines is largely consistent with that on Qwen2.5-Omni, with FastV-om being the weakest and DivPrune-om and DASH the strongest competitors.

\clearpage

\input{table/table_ovinci_main_all.tex}
%subsec+++++++++++++
\subsection{Full Results on OmniVinci-9B} \label{sec:appendix_omnivinci_main}
%subsec+++++++++++++
OmniVinci-9B adopts a different backbone architecture and training strategy from the Qwen-Omni series. \cref{table:ovinci_main_all} presents the full results, complementing~\cref{ssec:main_results}. \ourmethod achieves a mean score of 53.0 at $R{=}35\%$, surpassing the full-token baseline (52.9), and retains 95.3\% of the full-token performance at $R{=}10\%$. Consistent with the trends on Qwen-Omni, \ourmethod ranks first across all retention ratios, with its advantage growing as $R$ decreases. It is worth noting that the ``-om'' variants, \eg VisionZip-om and DivPrune-om, consistently outperform their visual-only counterparts on this model, suggesting that joint compression of both modalities is more effective on OmniVinci.

\clearpage

%subsec+++++++++++++
\subsection{Detailed Analysis of $\lambda$} \label{sec:appendix_hyperparameter}
%subsec+++++++++++++
The scale factor $\lambda$ controls the token budget split between the pre-LLM and inner-LLM stages. Specifically, the pre-LLM \trr is set to $\lambda\cdot R$ ($\lambda>1$), so a larger $\lambda$ retains more tokens at the pre-LLM stage and shifts more compression to the inner-LLM relevance-based selection.
\cref{fig:param_ablation} shows the overall trend of $\lambda$ \wrt the mean score on Qwen2.5-Omni at $R{=}35\%$, and \cref{table:ablation_param_ers} reports the per-benchmark breakdown. As $\lambda$ increases from 1.25 to 1.45, the mean score rises from 51.07 to a peak of 51.44 at $\lambda{=}1.40$, before slightly dropping to 51.37 at $\lambda{=}1.45$. We adopt $\lambda{=}1.40$ as the default, as it achieves the highest mean and yields a more balanced distribution across individual benchmarks.

\input{table/ablation/table_ablation_param_ers_sigma}

%subsec+++++++++++++
\subsection{Detailed Robustness to Block Boundaries} \label{sec:appendix_boundary_robustness}
%subsec+++++++++++++

\cref{fig:boundary_sensitivity} shows the change in mean score when shifting the block boundaries $L_s$ or $L_l$ by $\pm 2$ layers. \cref{table:ablation_layer_sensitivity} further provides the per-benchmark breakdown. Across all three om-LLMs, $\Delta$Mean stays within 0.21 for every shifted configuration, and individual benchmark scores fluctuate only marginally. This confirms that the performance of \ourmethod is not sensitive to the exact positions of the transition layers. As such, a simple model-agnostic fixed depth-proportion strategy, \ie placing the boundaries at fixed percentages of the LLM depth, suffices to achieve stable performance across different om-LLMs.

\input{table/ablation/table_ablation_layer_sensitivity}

%% file: table/table_qw3o_main_all.tex
\begin{table}[t!]
\caption{Comparison of different methods on Qwen3-Omni-30B-A3B-Instruct.}
\label{table:qw3o_main_all}

\centering
\setlength{\tabcolsep}{3.5pt} % 设置表格中列间距
\renewcommand{\arraystretch}{1.0} % 调整表格行高度

\resizebox{1.0\linewidth}{!}{
\begin{tabular}{@{}l| l c | cccccc c@{}}
\toprule
\textbf{Method} & $R$ (\textcolor{vcolor}{$R_v$}-\textcolor{acolor}{$R_a$}) & \textbf{TFLOPs$\downarrow$} & \textbf{\specialcell{World\\Sense}} & \textbf{\specialcell{Daily\\Omni}} & \textbf{\specialcell{AVUT}} & \textbf{\specialcell{OmniVideo\\Bench}} & \textbf{\specialcell{Video\\MME}} & \textbf{\specialcell{LVOmni\\Bench}} & \textbf{Mean} \\
\midrule

Full tokens &
100 \var{100,100} & 47.2 &
54.3 & 71.8 & 74.9 & 40.3 & 73.1 & 38.0
& 58.7$_{\text{\;\scriptsize 100.0\%}}$ \\

\midrule

%%%%%%%%%%%%%%  35 Group  %%%%%%%%%%%%%%
Random &
35 \var{32,70} & 13.4 &
52.6 & 69.5 & 71.6 & 39.5 & 70.7 & 35.7
& 56.6$_{\text{\;\scriptsize 96.4\%}}$ \\

FastV &
35 \var{29,100} & 13.7 &
51.5 & 68.6 & 71.2 & 40.0 & 69.5 & 37.1
& 56.3$_{\text{\;\scriptsize 95.9\%}}$ \\

FastV-om &
35 \var{32,70} & 13.7 &
49.7 & 66.4 & 69.0 & 37.8 & 67.5 & 35.7
& 54.4$_{\text{\;\scriptsize 92.7\%}}$ \\

VisionZip &
35 \var{29,100} & 13.4 &
53.0 & \underline{71.4} & 72.7 & \underline{40.3} & 70.6 & 36.6
& 57.4$_{\text{\;\scriptsize 97.8\%}}$ \\

VisionZip-om &
35 \var{32,70} & 13.4 &
53.3 & 71.1 & 72.4 & 39.1 & 71.6 & \underline{37.5}
& 57.5$_{\text{\;\scriptsize 98.0\%}}$ \\

DivPrune &
35 \var{29,100} & 13.4 &
53.4 & 69.8 & \underline{73.2} & 39.6 & 70.7 & 36.6 &
57.2$_{\text{\;\scriptsize 97.4\%}}$ \\

DivPrune-om &
35 \var{32,70} & 13.4 &
53.2 & 70.4 & 72.9 & 39.5 & 71.5 & 36.6 &
57.4$_{\text{\;\scriptsize 97.8\%}}$ \\

DyCoke &
35 \var{29,100} & 13.4 &
52.9 & 68.6 & 72.0 & 40.1 & 70.3 & 36.6
& 56.8$_{\text{\;\scriptsize 96.8\%}}$ \\

OmniZip &
35 \var{32,70} & 13.4 &
\underline{53.8} & 70.9 & 73.0 & 40.1 & 71.6 & 36.8
& \underline{57.7}$_{\text{\;\scriptsize 98.3\%}}$ \\

DASH &
35 \var{32,70} & 13.4 &
53.2 & 70.2 & 73.0 & 39.2 & \underline{71.7} & 37.4
& 57.4$_{\text{\;\scriptsize 97.8\%}}$ \\

\rowcolor[HTML]{f0edff}
\ourmethod &
35 \var{32,70} & 13.3 &
\textbf{53.9} & \textbf{72.4} & \textbf{73.8} & \textbf{40.4} & \textbf{72.1} & \textbf{38.8}
& \textbf{58.6}$_{\text{\;\scriptsize 99.8\%}}$ \\

\midrule

%%%%%%%%%%%%%%  25 Group  %%%%%%%%%%%%%%
Random &
25 \var{22,60} & 9.4 &
51.3 & 68.1 & 70.3 & 39.0 & 70.4 & 35.6
& 55.8$_{\text{\;\scriptsize 95.1\%}}$ \\

FastV &
25 \var{18,100} & 9.8 &
49.5 & 67.0 & 68.8 & 39.7 & 66.1 & 35.7
& 54.5$_{\text{\;\scriptsize 92.8\%}}$ \\

FastV-om &
25 \var{22,60} & 9.8 &
47.2 & 62.9 & 65.3 & 38.1 & 65.7 & 33.7
& 52.2$_{\text{\;\scriptsize 88.9\%}}$ \\

VisionZip &
25 \var{18,100} & 9.4 &
51.9 & 68.0 & 70.4 & 39.0 & 68.8 & 36.7
& 55.8$_{\text{\;\scriptsize 95.1\%}}$ \\

VisionZip-om &
25 \var{22,60} & 9.4 &
51.8 & \underline{69.8} & 72.0 & 39.8 & 71.2 & 36.4
& 56.8$_{\text{\;\scriptsize 96.8\%}}$ \\

DivPrune &
25 \var{18,100} & 9.4 &
52.1 & 69.5 & \underline{72.2} & 39.6 & 70.2 & 36.8 &
56.7$_{\text{\;\scriptsize 96.6\%}}$ \\

DivPrune-om &
25 \var{22,60} & 9.4 &
52.2 & 69.0 & 71.9 & 40.0 & 70.9 & 36.2 &
56.7$_{\text{\;\scriptsize 96.6\%}}$ \\

FastVID &
25 \var{18,100} & 9.1 &
51.4 & 68.3 & 69.7 & 40.2 & 68.0 & 36.5
& 55.7$_{\text{\;\scriptsize 94.9\%}}$ \\

\graytxt{OmniZip} &
\graytxt{25 \var{25, 25}} & \graytxt{8.5} &
\graytxt{50.7} & \graytxt{66.0} & \graytxt{69.3} & \graytxt{39.4} & \graytxt{70.1} & \graytxt{36.0} &
\graytxt{55.3$_{\text{\;\scriptsize 94.2\%}}$} \\

DASH &
25 \var{22,60} & 9.4 &
\underline{52.4} & 68.7 & 71.9 & \underline{40.3} & \underline{71.5} & \underline{36.8}
& \underline{56.9}$_{\text{\;\scriptsize 96.9\%}}$ \\

\rowcolor[HTML]{f0edff}
\ourmethod &
25 \var{22,60} & 9.0 &
\textbf{53.2} & \textbf{71.2} & \textbf{72.7} & \textbf{41.3} & \textbf{71.9} & \textbf{38.0}
& \textbf{58.1}$_{\text{\;\scriptsize 99.0\%}}$ \\

\midrule

%%%%%%%%%%%%%%  15 Group  %%%%%%%%%%%%%%
Random &
15 \var{12,50} & 5.8 &
49.8 & 64.6 & 65.7 & 36.8 & 67.9 & 33.9
& 53.1$_{\text{\;\scriptsize 90.5\%}}$ \\

FastV-om &
15 \var{12,50} & 6.3 &
44.6 & 58.4 & 58.3 & 35.0 & 61.1 & 34.4
& 48.6$_{\text{\;\scriptsize 82.8\%}}$ \\

VisionZip &
15 \var{8,100} & 5.8 &
49.8 & 65.4 & 67.4 & \underline{39.1} & 65.6 & 35.9
& 53.9$_{\text{\;\scriptsize 91.8\%}}$ \\

VisionZip-om &
15 \var{12,50} & 5.8 &
50.2 & 66.0 & 66.7 & 37.6 & 67.7 & \underline{36.2}
& 54.1$_{\text{\;\scriptsize 92.2\%}}$ \\

DivPrune &
15 \var{8,100} & 5.8 &
\underline{51.0} & \underline{67.8} & 69.5 & 38.0 & 66.8 & 35.2 &
54.7$_{\text{\;\scriptsize 93.2\%}}$ \\

DivPrune-om &
15 \var{12,50} & 5.8 &
50.7 & 66.3 & \underline{69.8} & 37.8 & 68.6 & 35.4 &
54.8$_{\text{\;\scriptsize 93.4\%}}$ \\

DASH &
15 \var{12,50} & 5.8 &
51.0 & 65.6 & 69.3 & 38.8 & \underline{69.3} & 35.8
& \underline{55.0}$_{\text{\;\scriptsize 93.7\%}}$ \\

\rowcolor[HTML]{f0edff}
\ourmethod &
15 \var{12,50} & 5.5 &
\textbf{52.1} & \textbf{70.0} & \textbf{71.3} & \textbf{40.7} & \textbf{70.9} & \textbf{36.3}
& \textbf{56.9}$_{\text{\;\scriptsize 96.9\%}}$ \\

\midrule

%%%%%%%%%%%%%%  10 Group  %%%%%%%%%%%%%%
Random &
10 \var{7,45} & 4.0 &
48.3 & 61.2 & 63.4 & 36.6 & 65.3 & 34.2
& 51.5$_{\text{\;\scriptsize 87.7\%}}$ \\

VisionZip-om &
10 \var{7,45} & 4.0 &
48.9 & 61.8 & 64.8 & 38.3 & 65.2 & 35.0
& 52.3$_{\text{\;\scriptsize 89.1\%}}$ \\

DivPrune-om &
10 \var{7,45} & 4.0 &
\underline{49.3} & \underline{64.0} & \underline{67.1} & \underline{38.7} & \underline{67.2} & \underline{35.3} &
\underline{53.6}$_{\text{\;\scriptsize 91.3\%}}$ \\

\rowcolor[HTML]{f0edff}
\ourmethod &
10 \var{7,45} & 3.9 &
\textbf{51.2} & \textbf{67.0} & \textbf{68.5} & \textbf{40.2} & \textbf{69.4} & \textbf{37.0}
& \textbf{55.6}$_{\text{\;\scriptsize 94.7\%}}$ \\

\bottomrule
\end{tabular}
}

\end{table}

%% file: table/table_ovinci_main_all.tex
\begin{table}[t!]
\caption{Comparison of different methods on OmniVinci-9B.}
\label{table:ovinci_main_all}

\centering
\setlength{\tabcolsep}{3.5pt} % 设置表格中列间距
\renewcommand{\arraystretch}{1.0} % 调整表格行高度

\resizebox{1.0\linewidth}{!}{
\begin{tabular}{@{}l| l c | cccccc c@{}}
\toprule
\textbf{Method} & $R$ (\textcolor{vcolor}{$R_v$}-\textcolor{acolor}{$R_a$}) & \textbf{TFLOPs$\downarrow$} & \textbf{\specialcell{World\\Sense}} & \textbf{\specialcell{Daily\\Omni}} & \textbf{\specialcell{AVUT}} & \textbf{\specialcell{OmniVideo\\Bench}} & \textbf{\specialcell{Video\\MME}} & \textbf{\specialcell{LVOmni\\Bench}} & \textbf{Mean} \\
\midrule

Full tokens &
100 \var{100,100} & 68.5 &
49.3 & 64.7 & 65.1 & 34.6 & 67.1 & 36.8
& 52.9$_{\text{\;\scriptsize 100.0\%}}$ \\
\midrule

%%%%%%%%%%%%%%  35 Group  %%%%%%%%%%%%%%
Random &
35 \var{32,65} & 29.7 &
47.0 & 60.7 & 60.0 & 33.6 & 67.3 & 34.6
& 50.5$_{\text{\;\scriptsize 95.5\%}}$ \\

FastV &
35 \var{28,100} & 29.8 &
44.9 & 63.6 & 59.4 & 33.5 & 63.6 & 35.0
& 50.0$_{\text{\;\scriptsize 94.5\%}}$ \\

FastV-om &
35 \var{32,65} & 29.8 &
47.2 & \underline{64.1} & 61.7 & 34.3 & 66.3 & 35.9
& 51.6$_{\text{\;\scriptsize 97.5\%}}$ \\

VisionZip &
35 \var{28,100} & 28.8 &
46.3 & 60.3 & 58.9 & 34.3 & 66.8 & 37.1
& 50.6$_{\text{\;\scriptsize 95.7\%}}$ \\

VisionZip-om &
35 \var{32,65} & 29.4 &
47.6 & 62.8 & \underline{62.3} & 34.3 & 69.0 & 37.2
& \underline{52.2}$_{\text{\;\scriptsize 98.7\%}}$ \\

DivPrune &
35 \var{28,100} & 29.7 &
47.2 & 62.8 & 61.9 & 34.0 & 66.3 & 34.5
& 51.1$_{\text{\;\scriptsize 96.6\%}}$ \\

DivPrune-om &
35 \var{32,65} & 29.4 &
47.8 & 62.5 & \underline{62.3} & 34.2 & 68.2 & 37.3
& 52.1$_{\text{\;\scriptsize 98.5\%}}$ \\

% FastVID &
% 35 \var{28,100} &  &
%  &  &  &  &  &
% &  \\

OmniZip &
35 \var{32,65} & 30.0 &
47.8 & 61.5 & 62.1 & \underline{34.6} & \underline{69.1} & 37.3
& 52.1$_{\text{\;\scriptsize 98.5\%}}$ \\

DASH &
35 \var{32,65} & 29.7 &
\underline{47.9} & 62.7 & 61.9 & 34.4 & 67.9 & \textbf{37.5}
& 52.1$_{\text{\;\scriptsize 98.5\%}}$ \\

\rowcolor[HTML]{f0edff}
\ourmethod &
35 \var{32,65} & 28.6 &
\textbf{48.4} & \textbf{64.2} & \textbf{63.8} & \textbf{34.9} & \textbf{69.4} & \underline{37.4}
& \textbf{53.0}$_{\text{\;\scriptsize 100.2\%}}$ \\

\midrule

%%%%%%%%%%%%%%  25 Group  %%%%%%%%%%%%%%
Random &
25 \var{22,55} & 23.1 &
46.5 & 57.3 & 57.7 & 34.1 & 66.0 & 34.8
& 49.4$_{\text{\;\scriptsize 93.4\%}}$ \\

FastV &
25 \var{17,100} & 23.3 &
43.6 & 61.2 & 55.5 & 32.3 & 61.8 & 34.7
& 48.2$_{\text{\;\scriptsize 91.1\%}}$ \\

FastV-om &
25 \var{22,55} & 23.3 &
45.5 & 61.9 & 59.2 & \underline{34.6} & 64.3 & 35.6
& 50.2$_{\text{\;\scriptsize 94.9\%}}$ \\

VisionZip &
25 \var{17,100} & 22.3 &
45.4 & 56.0 & 56.3 & 32.9 & 66.3 & 37.3
& 49.0$_{\text{\;\scriptsize 92.6\%}}$ \\

VisionZip-om &
25 \var{22,55} & 22.9 &
46.9 & 62.0 & 60.2 & 34.4 & \underline{68.7} & 37.9
& \underline{51.7}$_{\text{\;\scriptsize 97.7\%}}$ \\

DivPrune &
25 \var{17,100} & 23.1 &
44.8 & 58.3 & 59.9 & 34.2 & 65.8 & 36.9
& 50.0$_{\text{\;\scriptsize 94.5\%}}$ \\

DivPrune-om &
25 \var{22,55} & 23.4 &
47.1 & \underline{62.2} & 59.6 & 33.7 & 66.9 & 37.5
& 51.2$_{\text{\;\scriptsize 96.8\%}}$ \\

% FastVID &
% 25 \var{17,100} &  &
%  &  &  &  &  &
% &  \\

\graytxt{OmniZip} &
\graytxt{25 \var{25,25}} & \graytxt{15.1} &
\graytxt{44.8} & \graytxt{58.1} & \graytxt{55.4} & \graytxt{33.2} & \graytxt{66.6} & \graytxt{37.3}
& \graytxt{49.2$_{\text{\;\scriptsize 93.0\%}}$} \\

DASH &
25 \var{22,55} & 23.1 &
\underline{47.3} & 60.7 & \underline{60.6} & 34.4 & 66.5 & \textbf{38.2}
& 51.3$_{\text{\;\scriptsize 97.0\%}}$ \\

\rowcolor[HTML]{f0edff}
\ourmethod &
25 \var{22,55} & 22.0 &
\textbf{47.8} & \textbf{63.4} & \textbf{62.9} & \textbf{34.7} & \textbf{69.1} & \underline{38.1}
& \textbf{52.7}$_{\text{\;\scriptsize 99.6\%}}$ \\

\midrule

%%%%%%%%%%%%%%  15 Group  %%%%%%%%%%%%%%
Random &
15 \var{12,45} & 16.7 &
44.2 & 54.3 & 53.0 & 34.0 & 64.4 & 35.0
& 47.5$_{\text{\;\scriptsize 89.8\%}}$ \\

FastV-om &
15 \var{12,45} & 17.0 &
43.1 & \underline{58.7} & 53.3 & 32.0 & 61.6 & 36.2
& 47.5$_{\text{\;\scriptsize 89.8\%}}$ \\

VisionZip &
15 \var{6,100} & 16.4 &
43.1 & 53.3 & 54.3 & 33.1 & 64.0 & 37.6
& 47.6$_{\text{\;\scriptsize 90.0\%}}$ \\

VisionZip-om &
15 \var{12,45} & 16.5 &
45.3 & 57.6 & 58.1 & 34.5 & \underline{67.1} & 37.5
& \underline{50.0}$_{\text{\;\scriptsize 94.5\%}}$ \\

DivPrune &
15 \var{6,100} & 16.7 &
44.6 & 57.1 & 57.2 & 34.4 & 64.3 & 37.9
& 49.3$_{\text{\;\scriptsize 93.2\%}}$ \\

DivPrune-om &
15 \var{12,45} & 16.5 &
45.3 & 57.8 & \underline{58.7} & 34.5 & 66.0 & 37.9
& \underline{50.0}$_{\text{\;\scriptsize 94.5\%}}$ \\

DASH &
15 \var{12,45} & 16.7 &
\underline{45.7} & 57.1 & 56.3 & \textbf{35.0} & 66.2 & \textbf{38.2}
& 49.8$_{\text{\;\scriptsize 94.1\%}}$ \\

\rowcolor[HTML]{f0edff}
\ourmethod &
15 \var{12,45} & 16.2 &
\textbf{46.9} & \textbf{61.5} & \textbf{60.6} & \underline{34.7} & \textbf{68.2} & \underline{38.0}
& \textbf{51.7}$_{\text{\;\scriptsize 97.7\%}}$ \\

\midrule

%%%%%%%%%%%%%%  10 Group  %%%%%%%%%%%%%%
Random &
10 \var{7,40} & 13.6 &
42.9 & 50.2 & 52.7 & 32.9 & 63.3 & 35.3
& 46.2$_{\text{\;\scriptsize 87.3\%}}$ \\

VisionZip-om &
10 \var{7,40} & 13.7 &
\underline{44.6} & 54.5 & 55.7 & \underline{33.8} & 65.0 & \underline{37.8}
& 48.6$_{\text{\;\scriptsize 91.9\%}}$ \\

DivPrune-om &
10 \var{7,40} & 13.7 &
44.1 & \underline{55.8} & \underline{56.3} & 33.4 & \underline{65.4} & 37.6
& \underline{48.8}$_{\text{\;\scriptsize 92.2\%}}$ \\

\rowcolor[HTML]{f0edff}
\ourmethod &
10 \var{7,40} & 13.0 &
\textbf{46.5} & \textbf{59.4} & \textbf{58.0} & \textbf{34.3} & \textbf{66.1} & \textbf{38.0}
& \textbf{50.4}$_{\text{\;\scriptsize 95.3\%}}$ \\

\bottomrule
\end{tabular}
}
\end{table}

%% file: table/ablation/table_ablation_param_ers_sigma.tex
\begin{table}[h!]
\caption{Effect of the scale hyperparameter $\lambda$.
}
\label{table:ablation_param_ers}

\centering
\setlength{\tabcolsep}{3.5pt} % 设置表格中列间距
\renewcommand{\arraystretch}{1.05} % 调整表格行高度

\resizebox{1.0\linewidth}{!}{
\begin{tabular}{@{}l| cccccc c@{}}
\toprule
\textbf{Config} & \textbf{WorldSense} & \textbf{\specialcell{Daily-Omni}} & \textbf{AVUT} & \textbf{\specialcell{OmniVideoBench}} & \textbf{Video-MME} & \textbf{\specialcell{LVOmniBench}} & \textbf{Mean} \\
\midrule

$\lambda=1.25$ & 45.6 & 61.4 & 62.4 & 34.6 & 66.9 & 35.5 & 51.1 \\
$\lambda=1.30$ & 45.5 & 61.6 & 62.2 & 34.6 & 66.9 & 35.9 & 51.1 \\
$\lambda=1.35$ & 45.4 & 61.8 & 62.4 & 34.8 & 66.9 & 36.0 & 51.2 \\

\rowcolor[HTML]{f0edff}
$\lambda=1.40$ & 46.3 & 62.1 & 62.7 & 35.0 & 66.6 & 36.0 & 51.4 \\
$\lambda=1.45$ & 45.9 & 61.8 & 62.6 & 34.9 & 66.6 & 36.4 & 51.4 \\
\bottomrule
\end{tabular}
}

\end{table}

%% file: table/ablation/table_ablation_layer_sensitivity.tex
\begin{table}[h!]
% \caption{Sensitivity to layer boundaries at $R{=}35\%$. Shifting $L_s$ or $L_l$ by $\pm 2$ layers, with intermediate layers recomputed accordingly, changes the mean score only marginally across all three models.}
\caption{Robustness to block boundaries at $R{=}35\%$, with per-benchmark scores reported.}
\label{table:ablation_layer_sensitivity}

\centering
\setlength{\tabcolsep}{4.5pt}
\renewcommand{\arraystretch}{1.0}

\resizebox{1.0\linewidth}{!}{
\begin{tabular}{@{}l | c | cccccc c|c@{}}
\toprule
\textbf{Config} & \textbf{\specialcell{Transition Layer\\$L_s-L_{m_1}-L_{m_2}-L_l$}} & \textbf{\specialcell{World\\Sense}} & \textbf{\specialcell{Daily\\Omni}} & \textbf{\specialcell{AVUT}} & \textbf{\specialcell{OmniVideo\\Bench}} & \textbf{\specialcell{Video\\MME}} & \textbf{\specialcell{LVOmni\\Bench}} & \textbf{Mean} & \textbf{$\Delta$Mean} \\
\midrule

\rowcolor{gray!15}
\multicolumn{10}{@{}c@{}}{\textit{Qwen2.5-Omni-7B}} \\ [2pt]

\rowcolor[HTML]{f0edff}
\ourmethod & $15-17-20-23$ &
46.3 & 62.1 & 62.7 & 35.0 & 66.6 & 36.0
& 51.44 & -- \\

$L_s{-}2$ & $13-16-20-23$ &
45.7 & 62.6 & 62.5 & 34.9 & 66.7 & 35.9
& 51.36 & $-0.08$ \\

$L_l{-}2$ & $15-17-19-21$ &
46.1 & 61.9 & 63.0 & 34.9 & 66.8 & 35.7
& 51.40 & $-0.04$ \\

$L_s{+}2$ & $17-18-20-23$ &
46.3 & 62.1 & 62.9 & 35.2 & 66.9 & 36.0
& 51.55 & $+0.11$ \\

$L_l{+}2$ & $15-17-21-25$ &
46.1 & 62.0 & 62.9 & 35.1 & 66.5 & 36.0
& 51.42 & $-0.02$ \\

\midrule

\rowcolor{gray!15}
\multicolumn{10}{@{}c@{}}{\textit{OmniVinci-9B}} \\ [2pt]

\rowcolor[HTML]{f0edff}
\ourmethod & $15-17-20-23$ &
48.4 & 64.2 & 63.8 & 34.9 & 69.4 & 37.4
& 53.01 & -- \\

$L_s{-}2$ & $13-16-20-23$ &
48.3 & 64.4 & 63.4 & 35.3 & 69.1 & 37.4
& 52.98 & $-0.03$ \\

$L_l{-}2$ & $15-17-19-21$ &
48.5 & 64.3 & 63.6 & 35.4 & 69.2 & 36.8
& 52.97 & $-0.04$ \\

$L_s{+}2$ & $17-18-20-23$ &
48.4 & 64.7 & 63.7 & 35.1 & 69.3 & 36.9
& 52.99 & $-0.02$ \\

$L_l{+}2$ & $15-17-21-25$ &
48.4 & 63.8 & 63.7 & 35.4 & 69.4 & 37.4
& 53.03 & $+0.02$ \\

\midrule

\rowcolor{gray!15}
\multicolumn{10}{@{}c@{}}{\textit{Qwen3-Omni-30B}} \\ [2pt]

\rowcolor[HTML]{f0edff}
\ourmethod & $25-29-34-39$ &
53.9 & 72.4 & 73.8 & 40.4 & 72.1 & 38.8
& 58.55 & -- \\

$L_s{-}2$ & $23-28-34-39$ &
53.6 & 72.0 & 73.8 & 40.1 & 72.2 & 38.7
& 58.36 & $-0.19$ \\

$L_l{-}2$ & $25-29-33-37$ &
53.7 & 72.0 & 73.7 & 39.9 & 72.2 & 38.8
& 58.37 & $-0.18$ \\

$L_s{+}2$ & $27-30-34-39$ &
54.0 & 72.2 & 73.6 & 40.0 & 72.2 & 38.7
& 58.44 & $-0.11$ \\

$L_l{+}2$ & $25-29-35-41$ &
53.9 & 72.0 & 73.8 & 40.1 & 72.1 & 38.6
& 58.34 & $-0.21$ \\

\bottomrule
\end{tabular}
}
\end{table}

%% file: sec/appendix/3_exp_details.tex
\clearpage
%-----------------------------------------------------------
\section{Additional Experimental Details} \label{sec:app_exp_details}
%-----------------------------------------------------------

%subsec+++++++++++++
\subsection{Evaluation Benchmarks}
\label{app:benchmark_details}
%subsec+++++++++++++

The experiments are conducted on six widely used video understanding benchmarks, summarized in~\cref{table:benchmark}. They span diverse temporal scales and differ in their reliance on audio versus visual cues, enabling a thorough assessment of compression robustness. We describe each benchmark below.
\input{table/table_benchmark}

\textbf{WorldSense}~\citep{hong2025worldsense} comprises 1,662 synchronized audio-visual videos spanning 8 domains, with 3,172 expert-annotated multiple-choice questions across 26 cognitive tasks. Its core design principle is tight audio-visual coupling: every question requires jointly integrating visual and audio evidence, and removing either modality leads to a drastic accuracy drop. The average video duration is approximately 141 seconds.

\textbf{Daily-Omni}~\citep{zhou2025dailyomni} collects 684 real-world YouTube videos of 30 to 60 seconds with 1,197 multiple-choice questions across six task families. Its distinguishing feature is temporal precision: answering requires pinpointing the correspondence between audio events and visual actions along the timeline, not global semantic matching.

\textbf{AVUT}~\citep{Yang2025avut} targets audio-centric video understanding, defining eight task types that span audio content understanding and audio-visual alignment. An answer-permutation filter removes questions answerable from text alone. We use the \textit{AV-Human} partition, which contains 691 expert-annotated videos with 1,734 multiple-choice QA pairs across 18 audio-centric domains, with an average duration of approximately 69 seconds.

\textbf{OmniVideoBench}~\citep{li2025omnivideobench} comprises 628 videos ranging from several seconds to 30 minutes, with 1,000 manually annotated multiple-choice questions covering 13 task types. Each question is accompanied by a multi-step reasoning chain (5.68 steps on average), explicitly recording the modalities and evidence involved, making it well suited for diagnosing weak links in a model's reasoning pipeline.

\textbf{Video-MME}~\citep{fu2025videomme} contains 900 videos across 6 domains with durations from 11 seconds to 1 hour, divided into short, medium, and long tiers, yielding 2,700 human-annotated QA pairs. It supports subtitle and audio auxiliary inputs. We report results \emph{without subtitles} to exclude external textual cues and isolate the effect of token compression.

\textbf{LVOmniBench}~\citep{tao2026lvomnibench} is, to our knowledge, the only benchmark dedicated to ultra-long audio-visual understanding. It curates 275 videos of 10 to 90 minutes (34.5 minutes on average, 140 hours in total) with 1,014 multiple-choice questions. Its average duration is 6 to 20 times longer than that of existing omni-modal benchmarks, specifically stressing multi-modal information retention and temporal localization over extended sequences.

%subsec+++++++++++++
\subsection{Reproduction Details of \ourmethod}
\label{app:method_details}
%subsec+++++++++++++

\textbf{Implementation}.
For Qwen-Omni series, video frames are uniformly sampled at 2 FPS. Following \citep{tao2025omnizip},
each time window contains 288 visual tokens, along with 50 audio tokens for Qwen2.5-Omni-7B and 26 for Qwen3-Omni-30B.
For Qwen2.5-Omni-7B, the maximum number of input frames is set to 128 for WorldSense, Daily-Omni and AVUT, 256 for OmniVideoBench, and 768 for Video-MME and LVOmniBench. As for Qwen3-Omni-30B, due to its larger memory consumption, the maximum number of input frames is set to 128 for the first three benchmarks and 196 for the remaining three.
For OmniVinci-9B, we follow the official configuration with 128 uniformly sampled frames for WorldSense, Daily-Omni, and AVUT, and 256 for the remaining three long benchmarks, with each frame producing 256 visual tokens and audio tokenized at 25 tokens per second.
Unless otherwise specified, our hyperparameter setting is as follows: $\lambda{=}1.4$.
For a fair comparison, we evaluate each method with the same $R$, chosen from $\{35\%, 25\%, 15\%, 10\%\}$.
All experiments are conducted on NVIDIA A800 80GB GPUs using LMMs-Eval~\citep{zhang2025lmmseval}.

\textbf{Decay schedule configuration}.
We detail the configuration of \ourmethod used to reproduce our results. Its inner-LLM block-wise decay schedule (\cref{ssec:method_exp_schedule}) is specified by the pre-LLM scale factor $\lambda$, the four transition layers that delimit the shallow, middle, and late blocks, and the per-sub-block scale factor $\delta$.

The four transition layers $L_s$, $L_{m_1}$, $L_{m_2}$, and $L_l$ follow a \textit{model-agnostic fixed depth-proportion} strategy, placed at $50\%$, $60\%$, $70\%$, and $80\%$ of the LLM's total depth, respectively, \ie $L_x=\lfloor \mathrm{L}\cdot p \rfloor + 1$ with $p\in\{0.5,0.6,0.7,0.8\}$. The resulting transition layers for different om-LLMs are listed in~\cref{table:transition_layer}.

\input{table/table_transition_layer}

Given these boundaries, the scale factor $\delta$ is determined by the token budget. As described in~\cref{ssec:method_exp_schedule}, the overall \trr $R$ equals the layer-count weighted average of the per-block \trrs. Since no token selection occurs in the shallow block (\trr $= r_s = \lambda R$) and all non-textual tokens are removed in the late block (\trr $= 0$), only the shallow and middle blocks contribute:
\begin{equation}
\begin{aligned}
R &= \frac{1}{\mathrm{L}}\Big(\underbrace{\sum_{i=1}^{L_s}  r_s}_{\text{shallow}} + \underbrace{\sum_{i=L_{s}+1}^{L_{m_1}} r_{m_1}}_{\text{middle sub-block 1}} + \underbrace{\sum_{i=L_{m_1}+1}^{L_{m_2}} r_{m_2}}_{\text{middle sub-block 2}} + \underbrace{\sum_{i=L_{m_2}+1}^{L_{l}} r_{m_3}}_{\text{middle sub-block 3}} \Big) \\
&= \frac{1}{\mathrm{L}}\Big( L_s\, r_s + (L_{m_1}\!-\!L_s)\, r_{m_1} + (L_{m_2}\!-\!L_{m_1})\, r_{m_2} + (L_l\!-\!L_{m_2})\, r_{m_3} \Big) \\
&= \frac{1}{\mathrm{L}}\Big( L_l\, \lambda R + \underbrace{\big(L_s + 2\, L_{m_1} + 3\, L_{m_2} -6\, L_l\big)}_{\text{Constant}~C} \cdot \delta \Big).
\end{aligned}
    \label{eq:budget_constraint_expanded}
\end{equation}

Solving for $\delta$:
\begin{equation}
    \delta = \frac{(\mathrm{L} - L_l\lambda)\, R}{C}, \quad C = L_s + 2\, L_{m_1} + 3\, L_{m_2} - 6\, L_l.
    \label{eq:delta_solution}
\end{equation}

For the Qwen2.5-Omni-7B configuration ($\mathrm{L}$=28, $\lambda$=1.4), we have $C = -29$. With $R$=0.3, this yields $\delta \approx 0.04345$.

\textbf{Budget allocation procedure}.
As described in~\cref{ssec:method_mmwin}, the relevance scores $S_{t,v}$ and $S_{t,a}$ are obtained from the last text token's query-key attention to the visual and audio tokens within each window. 
Concretely, the per-head softmax-normalized attention weights are averaged across heads to yield per-token scores. These scores are then aggregated per window and normalized via a softmax over windows.
In practice, the continuous budgets from~\cref{eq:intra_window_score_split} are rounded to integers. Any rounding residual is redistributed proportionally to $S_t$ so that the combined total $\sum_{t} B_t = r_v N_v + r_a N_a$ is maintained, with each window's budget capped at its current token count. Within the middle block, each modality retains at least one token per window. 
Since each transition layer selects from the tokens retained by earlier layers, removed tokens are never restored and the capacity constraint is naturally satisfied.

%subsec+++++++++++++
\subsection{Reproduction Details of Compared Baselines}
\label{app:baseline_details}
%subsec+++++++++++++

\cref{table:trr} lists the per-modality retention ratios used in our experiments. As stated in~\cref{ssec:experimental_setup}, each 2-second window contains $n_v{=}288$ visual tokens and $n_a$ audio tokens ($n_a{=}50$ for Qwen2.5-Omni, $n_a{=}26$ for Qwen3-Omni). 
For OmniVinci-9B, since it samples a fixed number of frames rather than at a fixed frame rate, the visual token count per window varies with video duration. Taking WorldSense (average 140\,s, 128 frames) as reference, each window contains approximately $n_v{=}468$ visual tokens and $n_a{=}50$ audio tokens, so the per-modality ratios are approximate. 
The overall budget constraint is $R_v \cdot n_v + R_a \cdot n_a = R \cdot (n_v + n_a)$. Since $n_a < n_v$, audio tokens constitute a smaller portion of the total budget, making it natural to preserve them and apply selection only to visual tokens. We therefore evaluate two modes. Under the \emph{Audio-intact} mode, $R_a{=}100\%$ and $R_v$ is solved accordingly (``--'' indicates cases where the resulting $R_v$ is impractically low). Under the \emph{Both-selected} mode, the budget is allocated to both modalities proportionally.
In what follows, we detail how each baseline is adapted to the om-LLM setting and which mode it operates under:

\input{table/table_trr}

\begin{itemize}[wide] % wide 增加列表项之间的水平间距, nosep 取消列表项之间的额外垂直间距
    \item \textbf{FastV\footnote{\url{https://github.com/pkunlp-icler/FastV}.}~\citep{chen2024fastv} (ECCV 2024).}
    FastV prunes tokens at the $K$-th LLM layer using cross-modal attention scores, with a pruning ratio $r$. We follow the official setting with $K{=}2$. The original method targets only visual tokens, so we evaluate it under the \emph{Audio-intact} mode. We also extend it to both visual and audio tokens, yielding FastV-om under the \emph{Both-selected} mode. The pruning ratio $r$ is set to match $R_v$ and $R_a$ in~\cref{table:trr} for fair comparison.

    \item \textbf{VisionZip\footnote{\url{https://github.com/dvlab-research/VisionZip}, Apache 2.0 License.}~\citep{yang2025visionzip} (CVPR 2025).}
    VisionZip selects dominant tokens by encoder attention and merges the rest into contextual tokens. Since the original method operates at the encoder output, conflicting with pooling in Qwen-Omni, we apply compression after pooling instead~\citep{shenfastvid}. We retain dominant and contextual tokens at a ratio of $(R{-}0.05){:}0.05$ per frame. The original method also targets only visual tokens, so we evaluate it under the \emph{Audio-intact} mode. We further extend it to both modalities with window-level compression for audio tokens, yielding VisionZip-om under the \emph{Both-selected} mode.

    \item \textbf{DivPrune\footnote{\url{https://github.com/vbdi/divprune}, CC BY-NC 4.0 License.}~\citep{alvar2025divprune} (CVPR 2025).}
    DivPrune retains a maximally diverse subset via greedy Max-Min cosine distance selection. The original method operates on pre-projector embeddings, which leads to noticeable degradation on om-LLMs. We therefore apply selection on post-projector embeddings after pooling instead. Selection is performed per frame for visual tokens. The original method targets only visual tokens, so we evaluate it under the \emph{Audio-intact} mode. We further extend it to both modalities with window-level selection for audio tokens, yielding DivPrune-om under the \emph{Both-selected} mode.

    \item \textbf{DyCoke\footnote{\url{https://github.com/KD-TAO/DyCoke}, Apache 2.0 License.}~\citep{tao2025dycoke} (CVPR 2025).}
    DyCoke operates at both the prefill and decode stages. Following the evaluation protocol of~\citep{tao2025omnizip}, we use only its prefill-stage TTM module, which partitions the video into 4-frame groups, keeps the first frame in each group intact, and merges temporally redundant visual tokens in the remaining frames based on inter-frame similarity. Since one frame is always preserved in each 4-frame window, the minimum video-token retention ratio $R_v$ is 25\%. As TTM targets only video tokens, we evaluate it under the \emph{Audio-intact} mode, which limits the lowest feasible $R$ to 35\%.

    \item \textbf{FastVID\footnote{\url{https://github.com/LunarShen/FastVID}, MIT License.}~\citep{shenfastvid} (NeurIPS 2025).}
    FastVID prunes visual tokens via spatiotemporal DPC-kNN. It dynamically segments video tokens based on transition similarities, selects salient tokens per frame, and merges the remaining ones by spatiotemporal redundancy elimination. We follow the official hyperparameters: minimum segment count $c{=}8$, segment threshold $\tau{=}0.9$, salient token ratio $d{=}0.4$, anchor frame step $p{=}4$, and merging factor $\alpha{=}0.6$. As it targets only video tokens, we evaluate it under the \emph{Audio-intact} mode.

    \item \textbf{OmniZip\footnote{\url{https://github.com/KD-TAO/OmniZip}, Apache 2.0 License.}~\citep{tao2025omnizip} (CVPR 2026).}
    OmniZip derives per-time-group retention scores from audio saliency to guide dynamic video token pruning, combined with interleaved spatiotemporal compression. The official implementation limits the per-group pruning ratio to 75\%, resulting in a minimum $R_v$ of 25\%. 
    To avoid constructing the full global audio attention matrix (over 30\,GB GPU memory), the latest implementation uses chunked global attention on Qwen2.5-Omni (query chunked, key over the full audio sequence) and window-local attention on Qwen3-Omni.
    As OmniZip handles both modalities, we evaluate it under the \emph{Both-selected} mode. At $R{=}25\%$, $R_v$ cannot be reduced below 25\%, thus we set $R_a{=}25\%$ and \graytxt{gray out} this entry. In addition, since OmniZip's audio branch merges 5\% contextual tokens, we count both selected and merged tokens toward $R_a$ to ensure a fair budget comparison.

    \item \textbf{DASH\footnote{\url{https://github.com/laychou666/DASH}, Apache 2.0 License.}~\citep{li2025dash} (ECCV 2026).}
    DASH detects semantic boundaries from cosine-similarity drops between adjacent audio embeddings, replacing OmniZip's fixed temporal windows with variable-length segments and scoring tokens via a tri-signal fusion of boundary probability, uniqueness, and encoder attention.
    As an omni-modal compression method, we likewise evaluate it under the \emph{Both-selected} mode. In the official implementation, projected video segment boundaries are not aligned to complete frame windows, so tail tokens shorter than one frame are discarded, causing the actual $R_v$ to fall slightly below the configured value. We omit $R{=}10\%$ as the resulting $R_v$ is impractically low.
\end{itemize}

All methods are evaluated using LMMs-Eval\footnote{\url{https://github.com/EvolvingLMMs-Lab/lmms-eval}, Apache 2.0 License.}~\citep{zhang2025lmmseval} for consistency. The base omni-modal large language models used are Qwen2.5-Omni-7B\footnote{\url{https://huggingface.co/Qwen/Qwen2.5-Omni-7B}, Apache 2.0 License.}~\citep{xu2025qwen25omni}, OmniVinci-9B\footnote{\url{https://huggingface.co/nvidia/omnivinci}, Apache 2.0 License.}~\citep{ye2025omnivinci} and Qwen3-Omni-30B-A3B-Instruct\footnote{\url{https://huggingface.co/Qwen/Qwen3-Omni-30B-A3B-Instruct}, Apache 2.0 License.}~\citep{xu2025qwen3omni}.

%% file: table/table_benchmark.tex
\begin{table}[h!]
\caption{\textbf{Six benchmarks used in our experiments}.}
\label{table:benchmark}
\centering
\setlength{\tabcolsep}{5pt} % 设置表格中列间距
\renewcommand{\arraystretch}{1.1} % 调整表格行高度

\resizebox{0.65\linewidth}{!}{
\begin{tabular}{@{}l rr r@{}}
\toprule
\textbf{Benchmark} & \textbf{\#Videos} & \textbf{\#QA Pairs} & \textbf{Duration} (sec) \\
\midrule
WorldSense \citep{hong2025worldsense} & 1,662  & 3,172 & 140.7 \\
Daily-Omni \citep{zhou2025dailyomni} & 684 & 1,197 & 43.2 \\
AVUT \citep{Yang2025avut} & 691 & 1,734 & 68.6 \\
OmniVideoBench \citep{li2025omnivideobench}  & 628  & 1,000 & 409.0 \\
Video-MME \citep{fu2025videomme} & 900 & 2,700 & 1021.3 \\
LVOmniBench \citep{tao2026lvomnibench} & 275 & 1,014 & 2069.7 \\
\bottomrule
\end{tabular}
}
\end{table}

%% file: table/table_transition_layer.tex
\begin{table}[h!]
\caption{\textbf{Transition layers under the \textit{model-agnostic depth-proportion} strategy}.}
\label{table:transition_layer}
\centering
\setlength{\tabcolsep}{4pt} % 设置表格中列间距
\renewcommand{\arraystretch}{1.1} % 调整表格行高度

\resizebox{0.75\linewidth}{!}{
\begin{tabular}{@{}l @{${}={}$} l ccc@{}}
\toprule

\multicolumn{2}{@{}l}{\textbf{Transition Layer}} &
\specialcell{\textbf{Qwen2.5-Omni-7B} \\($\mathrm{L}{=}28$)} &
\specialcell{\textbf{OmniVinci-9B} \\($\mathrm{L}{=}28$)} &
\specialcell{\textbf{Qwen3-Omni-30B} \\($\mathrm{L}{=}48$)} \\
\midrule

$L_s$    & $\lfloor 0.5\times\mathrm{L}\rfloor+1$ & 15 & 15 & 25 \\
$L_{m_1}$ & $\lfloor 0.6\times\mathrm{L}\rfloor+1$ & 17 & 17 & 29 \\
$L_{m_2}$ & $\lfloor 0.7\times\mathrm{L}\rfloor+1$ & 20 & 20 & 34 \\
$L_l$    & $\lfloor 0.8\times\mathrm{L}\rfloor+1$ & 23 & 23 & 39 \\
\bottomrule
\end{tabular}
}
% \vspace{-15pt}
\end{table}

%% file: table/table_trr.tex
% TRR(R) sweep across backbones and evaluation splits (Appendix B.2).
\begin{table}[h!]
\caption{\textbf{Per-modality retention ratios (\varapp{$R_v$,$R_a$}) for each \trr ($R$)}. 
\emph{Audio-intact}: only visual tokens are selected ($R_a{=}100\%$). \emph{Both-selected}: token selection applied to both modalities. 
}
\label{table:trr}
\centering
\setlength{\tabcolsep}{3pt} % 设置表格中列间距
\renewcommand{\arraystretch}{1.1} % 调整表格行高度

\resizebox{0.9\linewidth}{!}{
\begin{tabular}{@{}l c cc c cc c cc@{}}
\toprule

\multirow{2}{*}{\trr ($R$)} & &
\multicolumn{2}{c}{\textbf{Qwen2.5-Omni-7B}} & &
\multicolumn{2}{c}{\textbf{OmniVinci-9B}} & &
\multicolumn{2}{c@{}}{\textbf{Qwen3-Omni-30B}} \\
\cmidrule{3-4}\cmidrule{6-7}\cmidrule{9-10}

& & \textit{Audio-intact} & \textit{Both-selected} & & \textit{Audio-intact} & \textit{Both-selected} & & \textit{Audio-intact} & \textit{Both-selected} \\
\midrule

35 & & \varapp{24,100} & \varapp{30,65} & & \varapp{28,100} & \varapp{32,65} & & \varapp{29,100} & \varapp{32,70} \\
25 & & \varapp{12,100} & \varapp{20,55} & & \varapp{17,100} & \varapp{22,55} & & \varapp{18,100} & \varapp{22,60} \\
15 & & -- & \varapp{10,45} & & \varapp{6,100} & \varapp{12,45} & & \varapp{8,100} & \varapp{12,50} \\
10 & & -- & \varapp{6,35} & & -- & \varapp{7,40} & & -- & \varapp{7,45} \\
\bottomrule
\end{tabular}
}
\end{table}

%% file: sec/appendix/5_discussion.tex
%-----------------------------------------------------------
\section{Limitations} \label{sec:app_discussion}
%-----------------------------------------------------------

Although our method \ourmethod allocates token budgets in an input-dependent manner via query relevance scores, the model-agnostic fixed depth-proportion decay schedule remains fixed for the three om-LLMs in our paper. Exploring sample-adaptive compression warrants further investigation.
Our current method operates after the encoders and does not reduce the computational cost of visual and audio encoding itself.
Moreover, \ourmethod targets single-turn QA over complete audio-visual inputs. Incremental budget allocation for streaming inputs and cross-turn evidence reuse in multi-turn interactions are directions worthy of future investigation.